\documentclass[10pt,twocolumn,letterpaper]{article}
\usepackage{iccv}
\usepackage{times}
\usepackage{epsfig}
\usepackage{graphicx}
\usepackage{amsmath}
\usepackage{amssymb}

\usepackage{graphicx}
\usepackage{amsmath}
\usepackage{amssymb}
\usepackage{subfigure} 
\usepackage{color}
\usepackage{colortbl}
\usepackage[table]{xcolor}
\usepackage{numprint}
\usepackage{multirow,array,varwidth}
\usepackage{booktabs}
\usepackage{makecell}
\usepackage{bbm}
\usepackage{dsfont}
\usepackage{xspace}
\usepackage{amsthm}
\usepackage{float}

\newcommand{\train}{{\small\texttt{train}\xspace}}
\newcommand{\val}{{\small\texttt{val}\xspace}}
\newcommand{\test}{{\small\texttt{test}\xspace}}

\newcommand{\longvise}{\textbf{Vis}ual \textbf{E}ngagement\xspace}
\newcommand{\vise}{VisE\xspace}

\newcommand{\cvpara}[1]{\vspace{0.05in}\noindent\textbf{#1}}
\newcommand{\cvlitepara}[1]{\vspace{0.07in}\noindent\textbf{#1}}

\newcommand{\R}[0]{\mathds{R}}

\DeclareMathOperator*{\argmin}{arg\,min}

\newcommand{\reloss}[0]{Reaction}
\newcommand{\cmloss}[0]{Comments}

\newcommand{\random}[0]{Random Init}
\newcommand{\insup}[0]{IN-Sup}
\newcommand{\moco}[0]{MoCo-v2}
\newcommand{\wsl}[0]{IG-940M-IN}

\newcommand{\visesm}[0]{VisE-1.2M}
\newcommand{\viselg}[0]{VisE-250M}

\newcommand{\virtex}[0]{VirTex}
\newcommand{\grid}[0]{VQAGrid}

\newcommand{\icmlmtfm}[0]{ICMLM\textsubscript{tfm}}
\newcommand{\icmlmatt}[0]{ICMLM\textsubscript{att-fc}}
\newcommand{\icmlm}[0]{ICMLM}
\newcommand{\clip}[0]{CLIP}

\newcommand{\imagenet}[0]{ImageNet}
\newcommand{\longcub}[0]{Caltech-UCSD Birds-200-2011}
\newcommand{\cub}[0]{CUB-200-2011}
\newcommand{\ube}[0]{UnbiasedEmotion}
\newcommand{\politics}[0]{Politics}
\newcommand{\hm}[0]{Hateful Memes}

\definecolor{green_im}{rgb}{0.0, 0.5, 0.0}
\newcommand{\meanstd}[2] { #1 \textsubscript{$\pm$  #2} }
\newcommand{\ttbf}[1]{\textbf{\texttt{#1}}}

\newcommand{\graycell}{\cellcolor{gray!15}}

\newcommand{\Drop}[1]{\textcolor{red}{\xspace\tiny{\bf $\downarrow$#1}}}
\newcommand{\Rise}[1]{\textcolor{green_im}{\xspace\tiny{\bf $\uparrow$#1}}}

\definecolor{dark_blue}{rgb}{10, 120, 180}

\definecolor{citecolor}{RGB}{0, 113, 188}

\usepackage[pagebackref=true,breaklinks=true,colorlinks,citecolor=citecolor,bookmarks=false]{hyperref}

\iccvfinalcopy 

\def\iccvPaperID{5398} 

\ificcvfinal\fi

\begin{document}

\title{Exploring Visual Engagement Signals for Representation Learning}

\author{
  Menglin Jia\thanks{Equal contribution.}~$^{1,2}$\hspace{10pt}
  Zuxuan Wu\footnotemark[1]~$^{2,3}$\hspace{10pt}
  Austin Reiter$^{2}$\hspace{10pt}
  Claire Cardie$^{1}$\hspace{10pt}
  Serge Belongie$^{1}$\hspace{10pt}
  Ser-Nam Lim$^{2}$ \\
$^{1}$Cornell University
\qquad $^{2}$Facebook AI
\qquad $^{3}$Fudan University
}

\maketitle
\ificcvfinal\thispagestyle{empty}\fi

\begin{abstract}
Visual engagement in social media platforms comprises interactions with photo posts including comments, shares, and likes. In this paper, we leverage such {\longvise} clues as supervisory signals for representation learning.  However, learning from engagement signals is non-trivial as it is not clear how to bridge the gap between low-level visual information and high-level social interactions. We present \textbf{{\vise}}, a weakly supervised learning approach, which maps social images to pseudo labels derived by clustered engagement signals. 
We then study how models trained in this way benefit subjective downstream computer vision tasks such as emotion recognition or political bias detection.
Through extensive studies, we empirically demonstrate the effectiveness of {\vise} across a diverse set of classification tasks beyond the scope of conventional recognition~\footnote{Project page: \url{https://github.com/KMnP/vise}}.
\end{abstract}

\section{Introduction}
\label{sec: intro}

People post photos on social media to invite engagement and seek connections. 
A photo of a cute dog can resonate with other dog lovers and trigger reactions such as the ``like'' or ``love'' button or comments including ``what an adorable dog'' and ``look at those blue eyes!''
The widely available interactions with the photos posted on social media, which we call \emph{visual engagement}, 
contain rich semantic descriptions (``dog'', ``blue eyes'') and are far less expensive to obtain than manual annotations in standard computer vision tasks, including coarse and fine-grained class labels~\cite{imagenet_cvpr09,WahCUB_200_2011,zhou2017places}, bounding boxes~\cite{lin_2014_coco,Gupta_2019_lvis}, and image captions~\cite{chen2015microsoft}.

\begin{figure}[t]
\centering
\includegraphics[width=\columnwidth]{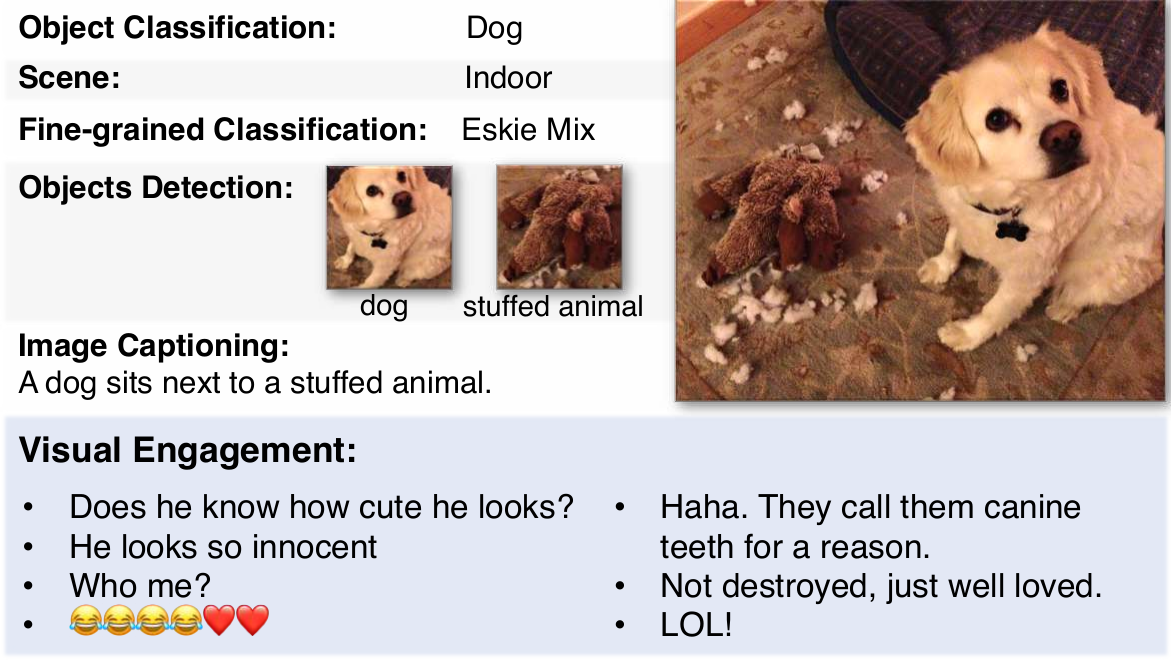}
\caption{
Visual engagement \vs other common supervisory signals. Given the same image, visual engagement provide semantically and contextually richer information than conventional recognition and captioning tasks.
}
\label{fig:teaser}
\end{figure}

More importantly, visual engagement, including comments, replies, likes, and shares, provides emotional and cultural context that goes beyond the image content therein. For example, the image in Fig.~\ref{fig:teaser} could be described in a standard captioning task as ``a dog sits next to a stuffed animal.''
The social media audience of this post may react to the cuteness of the dog, comment on the torn stuffed animal with whimsical responses, or initiate a conversation. 
The resulting textual descriptions depart from the \emph{exactly what it says on the tin} approach of standard image captioning tasks and express \emph{private states}~\cite{quirk2010comprehensive,wiebe2004sublearning}: opinions, emotions, and speculations, for example.
We argue that visual engagement can also serve as supervisory signals for representation learning and transfer well to subjective downstream computer vision tasks like emotion recognition or political bias classification.

Motivated by this observation, 
we propose to learn image representations from semantically and contextually rich {\longvise} signals (\textbf{{\vise}}).
We hypothesize that such learned representations, as a byproduct of mapping image content to human reactions, are able to infer private states expressed by images.
This is beneficial and could serve as a nice addition to current computer vision research which in general focuses on the objectively present factual information from images (\eg, ``this is a dog'' vs.~``what a cute dog!'').

Open-world visual engagement contains dense subjectivity clues, but is inherently noisy in nature.
How to properly leverage such signals for representation learning is a challenging, open question. Inspired by recent studies on feature learning from proxy tasks~\cite{GomezPatelCVPR2017self,caron2018deepcluster,xie2020noisy_student}, we cluster each type of visual engagement
and obtain cluster assignment indices for all the responses associated with a training image. These cluster assignments are used as supervisory cues. 
We then train a network from scratch to map images to cluster assignments in a multi-task fashion for representation learning, where each task is to predict the cluster index for that type of response. 
In this paper, we consider two forms of human responses: (1) comments and (2) reactions.
In the former case, we conduct the clustering on representations encoded by a textual model. Unlike most existing multi-modal methods that perform pre-training for both language and vision modules~\cite{desai2021virtex,radford2021learning} with hundreds of millions of parameters, we simply use an off-the-shelf encoder to embed comments, which is computationally efficient. We then evaluate representations learned from engagement signals on downstream tasks.

Our main contribution is to demonstrate that social media engagement can provide supervision for learning image representations that benefit
subjective downstream tasks. To this end, we explore \vise{} pre-trained on 250 million publicly available social media posts. Through extensive experiments, we show that in three downstream tasks related to private states detection, the learned \vise{} models can outperform the ImageNet-supervised counterpart by a substantial margin in some cases. These results highlight that \vise{} broadens the current representation learning paradigms, thereby narrowing the gap between machine and human intelligence.

\section{Related Work}\label{sec:related}

\cvpara{Visual representation learning}
Learning discriminative features for targeted datasets is a core problem for many computer vision research efforts. 
Feature learning on small datasets is particularly difficult since high-capacity deep neural networks suffer from overfitting when the amount of target training data is limited. 
One popular way to mitigate this issue is to pre-train on large-scale image datasets with manually curated class labels such as \imagenet{} and COCO~\cite{donahue2014decaf,dosovitskiy2021vit}.
However, training on \imagenet{} requires manually labeled data, which are expensive to obtain and hard to scale. 
This motivates a plethora of work investigating weakly-supervised~\cite{sun2017jft,wslimageseccv2018,kolesnikov2019large,kolensnikov2020bit,li2021mopro}, semi-supervised~\cite{yalniz2019billion,yan2020clusterfit,xie2020noisy_student} and self-supervised learning~\cite{caron2018deepcluster,zhuang2019localagg,donahue2019bigbigan,caron2020swav,2020byol,li2021pcl,zbontar2021barlowtwins}.
These methods tap into alternative forms of supervision, such as user-provided tags~\cite{yfcc100m,wslimageseccv2018} and hand-crafted pretext tasks
(\eg, inpainting~\cite{pathak2016inpainting}, colorization~\cite{zhang2016colorful,zhang2017split}, predicting jigsaw permutations~\cite{noroozi2016jigsaw}, rotations of inputs~\cite{gidaris2018rotnet}).
More recently, contrastive learning~\cite{gutmann2010nce,hadsell2006dimensionality} is used for feature learning by bringing images closer to their augmented versions than other samples in the training set~\cite{wu2018npid,ye2019e2e,he2019moco,misra2020pirl,chen2020simclr}. In this paper, we use visual engagement, which encompasses high-level semantics, as supervisory signals for representation learning.

\cvpara{Learning image representations using natural language}
There is growing interest in learning joint visual-language representations~\cite{frome2013devise,socher-etal-2014-grounded,karpathy2014,li2019vsrn,lu2019vilbert,chen2020vsepooling,chen2020uniter}.
Other studies convert language to discrete labels or continuous probability distributions, such as individual word, part-of-speech (POS) tags, clustering assignments of sentence features, and topic modeling~\cite{joulin2015flickr,li2017ngram,GomezPatelCVPR2017self,bulent2020icmlm}.
Some approaches learn visual representations from a pretext task that predicts the natural language captions from images~\cite{bulent2020icmlm,desai2021virtex,sariyildiz2020learning}.
Recently introduced methods including ConVIRT~\cite{zhang2020contrastive},CLIP~\cite{radford2021learning} and ALIGN~\cite{jia2021scaling} allow one to learn visual representations with contrastive objectives using image-text pairs.
These works use objective natural language, which describes and informs the content of the images, and mostly evaluate their methods on conventional recognition datasets.
We instead utilize the density of subjectivity clues in social media engagement and explore the transferability of the learnt representation to alternative downstream tasks.

\cvpara{Beyond conventional recognition}
Traditional computer vision tasks focus on the recognition of tangible properties of images, such as objects (both entry-level~\cite{imagenet_cvpr09,lin_2014_coco} and subordinate categories~\cite{WahCUB_200_2011,van_horn_inaturalist_2017,jia2020fashionpedia}) and scenes~\cite{zhou2017places}.
The research on representation learning mentioned above focuses on this type of task.
Relatively little attention has been paid to tasks that involve \emph{private states}~\cite{quirk2010comprehensive,wiebe2004sublearning} where subjectivity analysis is relevant.
This area includes (1) detecting cyberbullying and hate speech~\cite{hosseinmardi2016cyberbully,singh2017toward, gomez2020exploring,kiela2020hateful}, 
(2) identifying emotions~\cite{kosti2017emotion,bradley2005international,mollahosseini2017affectnet,panda2018ube,Wei_2020_CVPR}, 
(3) understanding rhetoric and intentions~\cite{joo2014visual,joo2015face,siddiquie2015exploiting,huang2016inferring,hussain2017automatic,ye2019interpreting,thomas2019predicting,kruk2019integrating,jia2020intentonomy}.
The present work aims to advance research in this area by learning effective features from high-level engagement signals.

\section{Approach}\label{sec:method}

With the aim of learning representations that capture the relationships between image content and human responses, we introduce a simple yet effective framework, \vise{}. 
It infers visual engagement signals from images (Sec.~\ref{subsec:methods_vise}). 
\vise{} is trained on large scale image-engagement pairs from a social media platform in a multi-task fashion, which will be described in Sec.~\ref{subsec:eng_types} and Sec.~\ref{subsec:method_details}.

\begin{figure}[t]
\centering
\includegraphics[width=\columnwidth]{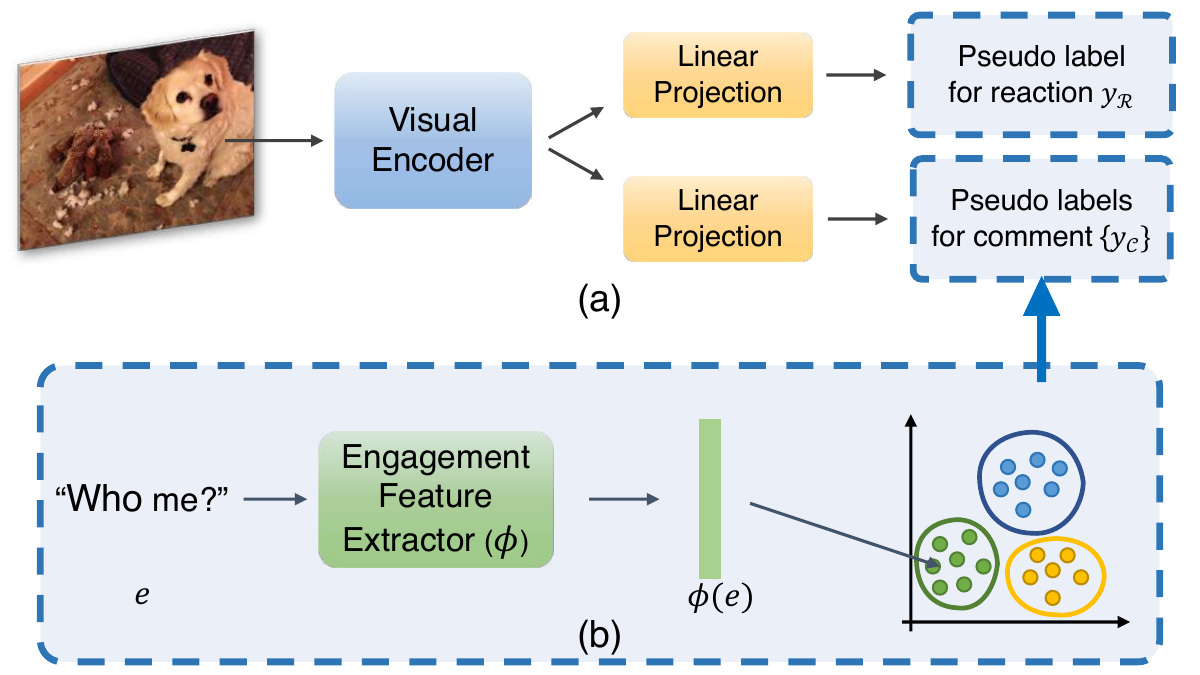}
\caption{Learning from visual engagement.
(a) We learn visual representations by predicting the pseudo-labels derived from a set of engagement signals including comments and raw reactions. (b) The pseudo-label is the clustered assignment index that is computed by first transforming the raw engagement ($e$) to a numerical representation $\phi(e)$.
See the main text for more details.
}
\label{fig:method}
\end{figure}

\subsection{From Engagement to Labels}
\label{subsec:methods_vise}
How people react to images is more telling than the content itself.
In this work, we propose to predict raw social engagement signals
by converting them into a bag-of-word multi-label multi-task classification task. Fig.~\ref{fig:method} illustrates the pipeline we use in this work.

More formally, let $(x, \{e\})$ be an image and a set of corresponding engagement clues (\eg, comments, replies, likes, \etc). 
Let $\phi$ be a general engagement feature extractor that transforms $e$ into a numerical representation $\phi(e) \in \R^{D}$. 
We describe the proposed method to preprocess and obtain pseudo-label for $e$ below. 

\paragraph{Step 1 (Cluster Generation)}
We first collect $e$ from a randomly sampled subset of the entire image-engagement pairs and generate $k$ clusters using $K$-means algorithm. This portion of the dataset will not be used during pre-training.

\paragraph{Step 2 (Label Creation)} Given the $i^{th}$ example from the unprocessed subset, $(x_i, \{e\}_i)$, we obtain a set of features $\{\phi(e) \}_i$ from the engagement set of this example, using the same function $\phi$. Next we collect the resulting cluster assignments indices as a set of labels $\{y_e\}$ for this image $x_i$. Fig.~\ref{fig:method}(a) summarizes this step.

\subsection{Engagement Types}\label{subsec:eng_types}
We use textual comments ($\mathcal{C}$) and raw reactions ($\mathcal{R}$) from publicly available Facebook posts as a first step to investigate visual engagement signals for representation learning.

\cvpara{Comments} Comments are direct human responses from image posts. Maximal 100 comments are randomly sampled from a post. We use the bag-of-word approach with the term frequency-inverse document frequency (TF-IDF) weighted document embeddings~\cite{salton1986ir} as comment feature extractor $\phi_{\mathcal{C}}$.
The cluster assignments derived from the comment set are used as multi-label classification targets, one label for each sampled comment associated with this specific image.

\cvpara{Raw reactions} Reactions are encoded as a normalized distribution over the five reaction buttons (``haha'', ``sorry'', ``angry'', ``wow'', ``love''). More specifically, we count the total occurrences of the 5 reaction buttons for each image post and normalize (L2) them to account for differences in followers and popularity of the post. 
Each post is mapped to a single cluster centroid index.

\subsection{Training \vise{} Models}\label{subsec:method_details}

\cvpara{\vise{} learning objective} \vise{} is trained in a multi-task learning framework by minimizing the following loss function:
\begin{align}
    \argmin_{W} \mathbb{E}_{(x_i, \{y_e\}_i)\sim\mathcal{D}} \,\, & \mathcal{L_C}( f(x_i), \{y_{\mathcal{C}}\}_i; W)  \nonumber \\
      +\, & \mathcal{L_R}( f(x_i), \{y_{\mathcal{R}}\}_i; W),
\end{align}
where $\mathcal{L_C}$ is a cross-entropy loss with soft probability~\cite{wslimageseccv2018}, $\mathcal{L_R}$ is the standard cross-entropy loss, $f$ represents an image feature encoder parameterized by $W$. $\mathcal{D}$ represents the training data which the image-engagement pairs are sampled from. In addition, $\{y_{\mathcal{C}}\}$ and $\{y_{\mathcal{R}}\}$ denote the pseudo labels of comments and reactions, respectively.

\cvpara{Pre-training details}
In our experiments, we train convolutional networks for engagement prediction. 
For comparison with previous work, the CNN model architectures we use are: ResNet-50, ResNet-101~\cite{he2016deep}, and ResNeXt-101 $32 \times 16$d~\cite{xie2017aggregated}.
\vise{} model with the largest backbone, ResNeXt-101 $32 \times 16$d, took around 10 days to train on 32 NVIDIA V100 GPUs with 250 million images with minibatch size of 1536.
Other pre-training details are included in the Appendix~\ref{supsec:detail}.

\section{Experiments}
\label{sec:exp}
We evaluate the effectiveness of feature representations learned by \vise for a wide range of downstream tasks. 
In our experiment, we aim to show that \emph{image representations learned from engagement signals are beneficial for image classification that beyond the scope of conventional recognition tasks}.
We begin by describing our experimental setups, including 
a comparison of alternative representation learning methods (Sec.~\ref{subsec:expsetup_method}), implementation details (Sec.~\ref{subsec:imp_details}), and a summary of evaluated downstream tasks (Sec.~\ref{subsec:expsetup_task}). 
Finally we present results and discussion in Sec.~\ref{subsec:expmain} and Sec.~\ref{subsec:cub_exp}.

\subsection{Compared methods}
\label{subsec:expsetup_method}

\begin{table*}[t]  
\scriptsize
\begin{center}
\resizebox{0.98\textwidth}{!}{%
\begin{tabular}{l l l l l l l c}
\Xhline{0.7pt}\noalign{\smallskip}
&\textbf{Method}  
&\textbf{Input Type} 
&\textbf{Annotation Type} &\textbf{Noisy Labels}
&\textbf{Pre-trained Data} &\textbf{Data Size} 
&\textbf{Model} \\  
\Xhline{0.7pt}\noalign{\smallskip}
&\random{} &- &- &- &train from scratch &-    \\
\hline\noalign{\smallskip}

\multirow{4}{*}{\ttbf{(1)}} &\insup{} 
&images &object labels  &
&ImageNet~\cite{imagenet_cvpr09} &1.28M 
&CNN  \\

&\wsl{}~\cite{wslimageseccv2018} 
&images &hashtags + labels &\checkmark 
&IG~\cite{wslimageseccv2018} + ImageNet~\cite{imagenet_cvpr09} &940M + 1.28M 
&CNN \\

&\moco{}~\cite{chen2020mocov2} 
&image pairs & - & 
&ImageNet~\cite{imagenet_cvpr09} &83.9B$^{\star}$
&CNN \\

&\grid{}~\cite{jiang2020defense} 
&images  &object + attribute &  
&VisualGenome~\cite{krishna2017visual} &103k 
&Faster-RCNN \\

\hline\noalign{\smallskip}


\multirow{3}{*}{\ttbf{(2)}} 
&\virtex{}~\cite{desai2021virtex} 
&images  &captions &
&COCO-caption~\cite{chen2015microsoft}  &118k 
&CNN + Transformer
\\

&\icmlm{}~\cite{bulent2020icmlm} 
&images + captions &masked token from captions  & 
&COCO-caption~\cite{chen2015microsoft}  &118k 
&CNN + Transformer \\

&\clip{}~\cite{radford2021learning} 
&images + text & -&
&WebImageText~\cite{radford2021learning}  &13.1T$^\star$
&CNN + Transformer \\
\hline\noalign{\smallskip}

& \multirow{2}{*}{\ttbf{ours}} 
&\multirow{2}{*}{images} 
&\multirow{2}{*}{pseudo labels }   & \multirow{2}{*}{\checkmark}
&\visesm &1.23M 
&\multirow{2}{*}{CNN} 
\\
&
& & &
&\viselg &250M 
&
\\
\Xhline{0.7pt}\noalign{\smallskip}
\end{tabular}
}
\caption{\vise{} \vs alternative methods compared. The seven representation learning approaches are grouped into: 
\ttbf{(1)} Uni-modal pre-training: similar to \vise{}, these methods use visual encoders only during pre-training;
\ttbf{(2)} Cross-modal pre-training: like \vise{}, these approaches learn from natural language as inputs or supervisory signals. All three architectures involve a Transformer-based model~\cite{vaswani2017attention} for the textual module. 
$^\star$We include the negative image-image / image-text pairs when counting the total data size for \moco{} and \clip{}. See Appendices for more details.
We also acknowledge that the effective data size is also affected by other factors such as data augmentations in other approaches. For simplicity, we use the actual dataset size for non-contrastive learning methods.}
\label{tab:methods}
\end{center}
\end{table*}

To train \vise{}, we collect a total of 270 million public image posts from a social media platform with 20 million used for cluster generation (see Sec.~\ref{subsec:methods_vise}).
To facilitate a fair comparisons with the ImageNet-supervised method, we also randomly sample 1.23 million images for pre-training.
We compare \vise{} pre-trained on 1.23 million (\visesm{}) and 250 million data (\viselg{}) with other feature representation learning methods. 

\cvpara{Uni-modal learning methods} We first compare \vise trained with pseudo labels derived from clustering assignments with networks that are trained with a pre-defined set of object labels:
\begin{itemize}
    \item ImageNet-supervised (\insup{}): the image encoder is pre-trained on ILSVRC 2012~\cite{imagenet_cvpr09} {\train} split (1.28M images)\footnote{Pre-trained models are from \href{https://github.com/pytorch/vision}{torchvision package} for ResNet-50/110.}. The dataset has 1000 classes, which is based on the concepts in WordNet~\cite{miller1995wordnet}.

    \item IG-imagenet (\wsl{})~\cite{wslimageseccv2018}: the visual encoder is pre-trained on 940 million public images with 1.5K hashtags in weakly-supervised fashion; the encoder is further fine-tuned on ImageNet dataset.

     \item \grid{}~\cite{jiang2020defense}: a pre-training method primarily for visual question answering and image captioning tasks. 
    It learns visual representations by training a Faster-RCNN~\cite{ren2015faster} on the Visual Genome dataset~\cite{krishna2017visual} which has 1600 object categories and 400 attributes. 
    We use the outputs from the last bottleneck block of the ResNet-50 as the pre-trained image representations.

    \item \moco{}~\cite{he2020momentum}: a self-supervised contrastive method using a momentum-based encoder and a memory queue trained on ImageNet. Given an image sample, it is trained to be closer to its randomly augmented version on a hypersphere than other samples in the dataset. We use the improved version~\cite{chen2020mocov2} that is trained with 800 epochs. Note that this method uses image as supervision labels instead of the ImageNet class labels.
\end{itemize}

\cvpara{Cross-modality pre-training}
Learning methods that use natural language as supervisory signals are also considered:
\begin{itemize}  
    \item \icmlm{}~\cite{bulent2020icmlm}: it  uses 118K image-text pairs from COCO-captions~\cite{chen2015microsoft} and uses masked
language modeling to learn visual representations from text. We include two versions of this method, \icmlmtfm{} and \icmlmatt{}, which respectively use a transformer and an attention-based mechanism for joint fine-tuning.

    \item \virtex~\cite{desai2021virtex}: this method also pre-trains on COCO-captions but on a different task: generating captions based on images. 
    
     \item \clip{}~\cite{radford2021learning}: Contrastive Language-Image Pre-training method (\clip{}) utilizes an image encoder and a text encoder to predict which images are paired with which textual descriptions in a large-scale dataset of 400M image-text pairs. 
\end{itemize}
It is worth pointing out all of these approaches train a text encoder to learn better natural language representations during pre-training stage, while \vise{} simply uses an off-the-shelf text encoder to compute representations for clustering purposes. We expect better performance of \vise{} if these textual representations are further fine-tuned~\cite{caron2018deepcluster} as these aforementioned methods.

We also report results of ``\random{}'', where no pre-trained features are used. 
Table~\ref{tab:methods} summarizes the differences between \vise{} and all of the baseline methods used in the experiment.

\subsection{Evaluation Protocols and Details} 
\label{subsec:imp_details}
We adopt two common protocols for evaluating the effectiveness of feature representations~\cite{wslimageseccv2018,goyal2019scaling,misra2020self,he2020momentum}:
(1) Linear evaluation: all pre-trained models are used as visual feature extractors, where the weights of the image encoders are fixed. This protocol is preferred for applications where computational resources or training data for target tasks are limited. The test performance indicates how effective the learned representations are for specific tasks.
(2) Fine-tuning: the parameters of the pre-trained image encoders are used as an advanced weight initialization method; these encoders are fine-tuned in an end-to-end manner for downstream tasks. Prior studies~\cite{girshick2014rich,ren2015faster} have shown that the latter protocol outperforms the linear evaluation approach due to its flexibility and adaptability to a wider range of downstream tasks.

See the Appendix~\ref{supsec:detail} for more implementation details, including a full list of hyperparameters used (batch sizes, learning rates, decay schedules, \etc) and sensitivity to hyperparameters for both linear and fine-tuned experiments.

\subsection{Downstream Tasks}
\label{subsec:expsetup_task}
We evaluate these visual representation methods on four downstream tasks, including sentiment classification, political bias, hate speech detection and fine-grained bird species classification.

\cvlitepara{\ube{} Dataset} 
This dataset~\cite{panda2018ube} contains 3045 images annotated into six emotional categories. To reduce object biases in the dataset, different emotion labels contain the same set of objects/scenes. Since there is no official split of this dataset, we random split the images into {\train} (70\%), {\val} (10\%), {\test} (20\%) set five times and report mean and standard deviation of the resulting accuracy.

\cvlitepara{\politics{}}
The task of this dataset~\cite{NEURIPS2019_politics} is to predict the political leaning (left and right) of images from news media. This dataset contains 749,932 images in total. Since only \train{} and \test set are publicly available, we randomly split the training set into \train{} (90\%) and \val{} (10\%) and report accuracy scores.

\cvlitepara{\hm{}}
\hm{} dataset~\cite{kiela2020hateful} contains multimodal memes including images and text. The task is to detect each meme is hate speech or not. We use the data from Phase 1 of the Hateful Memes challenge\footnote{\href{https://www.drivendata.org/competitions/64/hateful-memes/?fbclid=IwAR2LomJUKmStphTvTeMSuWFggaZDD3UO4gaJE0M9QXsT1ksB2SsXx1HVVic}{Hateful Memes: Phase 1 Challenge}}, which has 8500 training and 500 validation data. 
We obtain the sentence embeddings from a pre-trained RoBERTa model~\cite{liu2019roberta}, and concatenate the text features and image features together before linear evaluation that map the feature to the label space.
We report the macro averaged ROC AUC score and accuracy score on the {\val} set.

\cvlitepara{\longcub{} (\cub{})}
In addition to the above subjective classification tasks, we also evaluate our approach on standard image classification tasks. To this end, we use the \cub{} dataset.
\cub{}~\cite{WahCUB_200_2011} has a total of 11,788 images allocated over 200 (mostly North American) bird species. It is a benchmarking dataset for subordinate categorization. We train on the publicly available \train{} set and report top-1 accuracy on the \val{} set.

\begin{figure*}[h!]
\centering
\subfigure[Linear evaluation: \ube{}.]{
    \includegraphics[scale=0.4]{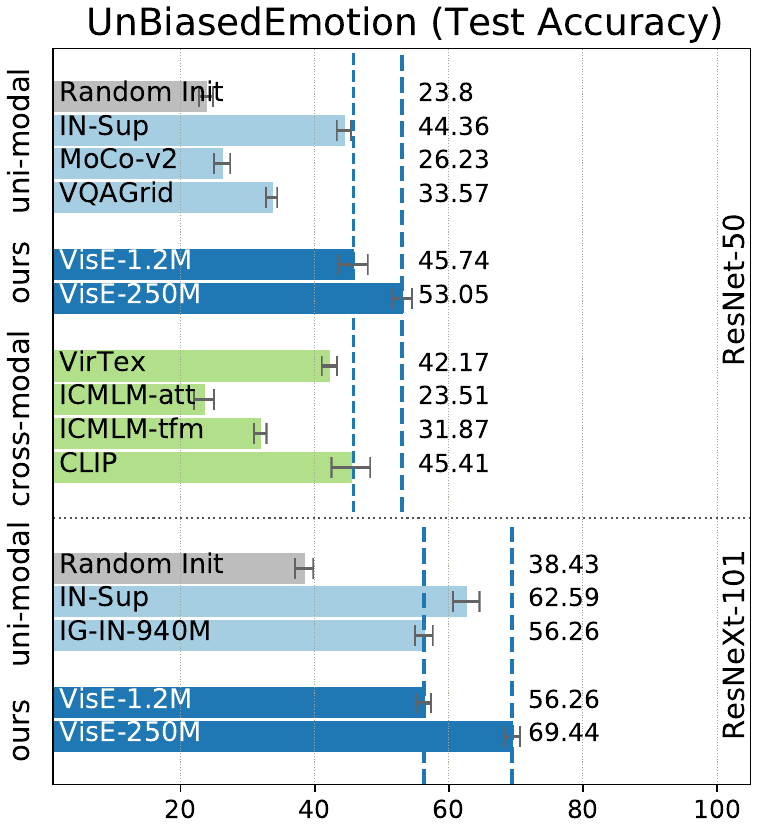}
    \label{fig:linear_ube}
}
\hfill
\subfigure[Linear evaluation: \politics{}.]{
    \includegraphics[scale=0.4]{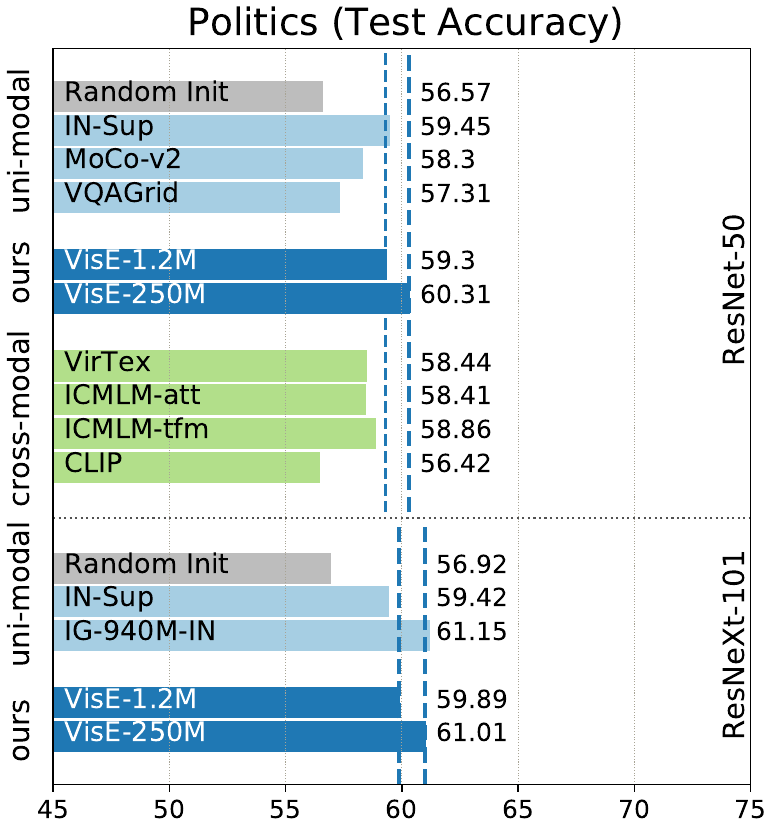}
    \label{fig:linear_politics}
}
\hfill
\subfigure[Linear evaluation: \hm{}.]{
    \includegraphics[scale=0.4]{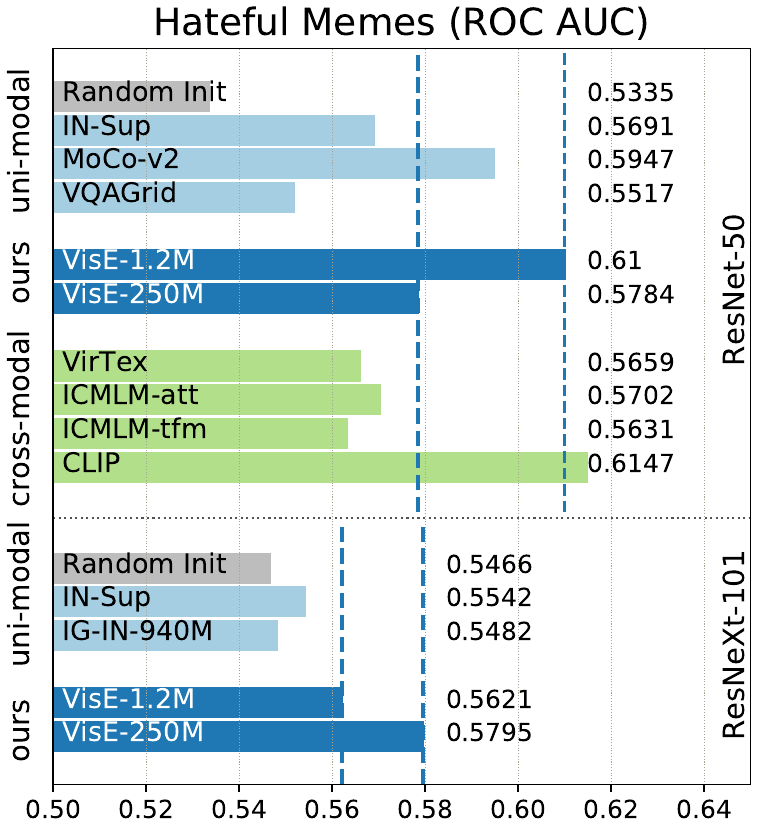}
    \label{fig:linear_hm}
}

\subfigure[Fine-tuned: \ube{}.]{
    \includegraphics[scale=0.4]{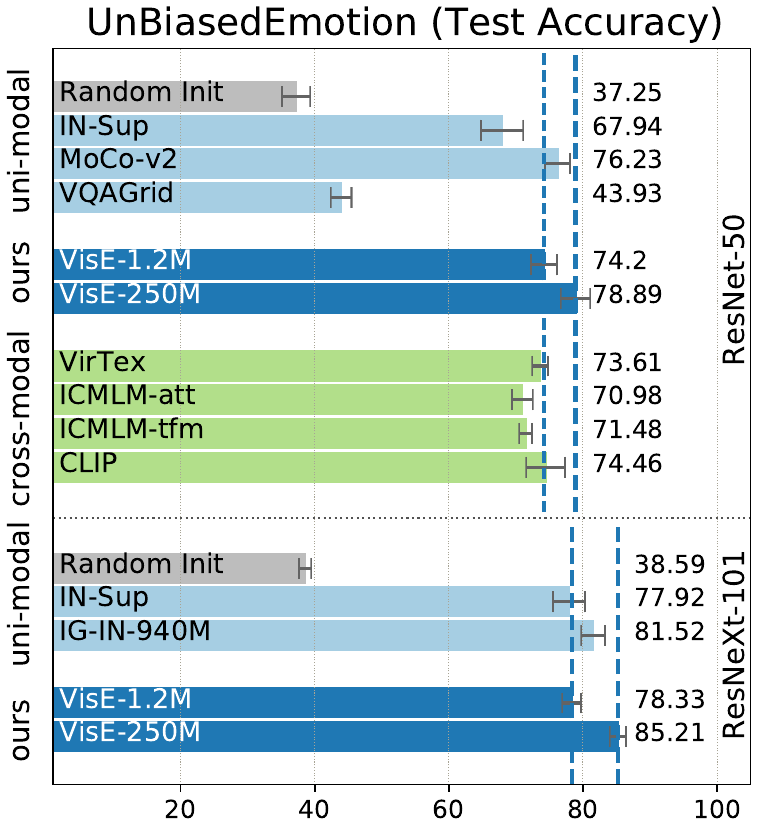}
    \label{fig:ft_ube}
}
\hfill
\subfigure[Fine-tuned: \politics{}.]{
    \includegraphics[scale=0.4]{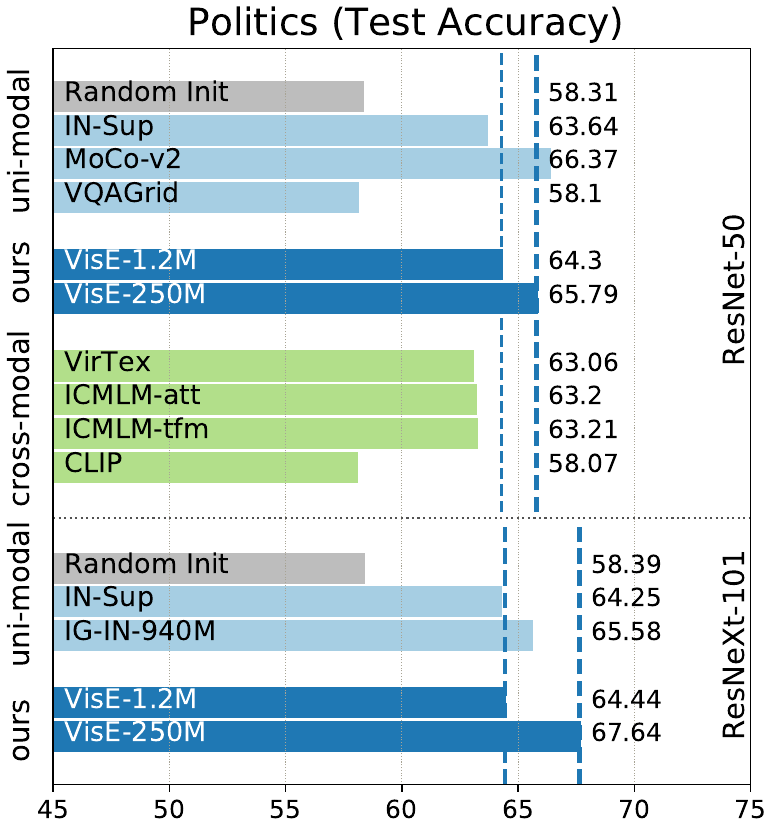}
    \label{fig:ft_politics}
}
\hfill
\subfigure[Fine-tuned: \hm{}.]{
    \includegraphics[scale=0.4]{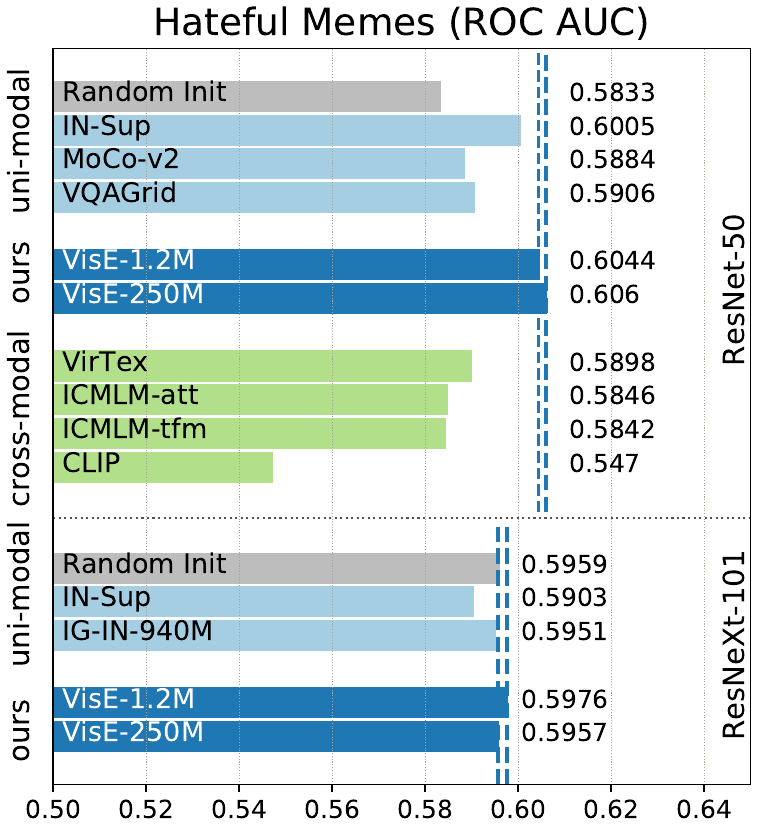}
    \label{fig:ft_hm}
}
\caption{Linear evaluation (\emph{top}) and fine-tuned (\emph{bottom}) results among \vise{}, uni-modal (ResNet-50), cross-modal (ResNet-50), and uni-modal (ResNeXt-101 32 $\times$ 16d) methods for all three datasets. For easy comparison, two blue dashed lines represent the performance of \visesm{} and \viselg{} respectively. 
}
\vspace{-0.3cm}
\label{fig:main}
\end{figure*}

\subsection{\vise{} for Subjective Recognition Tasks}
\label{subsec:expmain}
Following the two protocols described in Sec.~\ref{subsec:imp_details}, we compare the transfer-learning capabilities of \vise{} with other baseline approaches for three subjective tasks. 
Fig.~\ref{fig:main} presents the results for the linear evaluation and fine-tuned protocols, respectively.

\paragraph{\vise{} \vs other uni-modal methods}
From Fig.~\ref{fig:main}, we can see that:
(1) \vise{} is consistently better than other image-encoder only methods on three datasets over two visual backbone choices, except for \moco{} in \politics{}.
Even pre-trained with similar amounts of data (1.28M \vs 1.23M), \visesm{} still achieves better performance across all three tasks with a ResNet-50 backbone than models trained with labels from In-Sup.
(2) \viselg{} substantially outperforms \wsl{}, a method trained with a substantially larger amount of pre-trained data (950M \vs 250M).
(3) 
\moco{}, a self-supervised approach that does not require object category annotations, yields the best accuracy scores on \politics{} among approaches with ResNet-50 backbone. 
This also highlights the limitation of using object labels as pre-training supervision for such subjective tasks.

\cvpara{\vise{} \vs other methods that learn from language}
Under both protocols, \vise{} offers better or comparable results than other methods that leverage textual information during pre-training.
\vise{} achieves consistently better performance across all three tasks. This suggests that features learned with visual engagement signals are more suitable for subjective downstream tasks.
We also observe that all other four visual-language approaches obtain better results than \insup{} on \ube{} dataset when fine-tuned, but they are worse than \insup{} with linear evaluation. Such discrepancy might be caused by the scale of the dataset, since \ube{} is the smallest among the other tasks.

\subsection{\vise{} for Standard Recognition Tasks}
\label{subsec:cub_exp}
We compare the effectiveness of features learned from visual engagement signals with those trained using ImageNet labels on \cub{}. 
This task extends the general object classification from ImageNet and focuses on distinguishing fine-grained differences among 200 bird species. 
Moreover, 59 out of 1000 classes in ImageNet are already bird categories, including overlapping definitions with \cub{}~\cite{Horn2015}.
Thus Image-Net based approaches should transfer to this task better than \vise{}.
Table~\ref{tab:cub-results} shows the results of both linear evaluation and fine-tuned transfer protocols.

Indeed, \insup{} and \wsl{} achieve decent accuracy scores using linear classifiers alone to map the features to 200 bird species, outperforming \vise{} by a large margin.
This is foreseeable since visual engagement signals do not necessarily contain object information. 
It is understandable that \vise{} features are not as transferable to this task as models trained on ImageNet. 

When fine-tuning is performed, \vise with ResNet-50 have comparable or better performance than \insup{}. This highlights that fine-tuning the whole network can sometimes compensate for the inflexibility of learned features, which is in line with discussions in~\cite{radford2021learning}.

\begin{table}
\scriptsize
\begin{center}
\resizebox{0.9\columnwidth}{!}{%
\begin{tabular}{ l l  l l}
\Xhline{0.7pt}\noalign{\smallskip}

\textbf{Backbone} &\textbf{Method} &\textbf{Linear} &\textbf{Fine-tuned}
\\
\Xhline{0.7pt}\noalign{\smallskip}

\multirow{5}{*}{ResNet-50} &Random Init  
&3.42   &63.39
\\
&\insup{}
&\textbf{62.70} &72.15 
\\
\cline{2-4}\noalign{\smallskip}
&\visesm{}
&9.95 &72.65
\\
&\viselg{}
&9.92     &\textbf{76.58}
\\
\hline\noalign{\smallskip}
\multirow{5}{*}{\shortstack[l]{ResNeXt-101\\$32 \times 16$d}} &Random Init  
&7.04   &62.80 
\\
&\insup{}
&65.16   &84.21
\\
&\wsl{}~\cite{wslimageseccv2018} 
&\textbf{72.69}   &\textbf{85.28}
\\	
\cline{2-4}\noalign{\smallskip}
&\visesm{}
&9.76   &73.73 
\\	
&\viselg{}
&10.93   &79.54 
\\
\Xhline{0.7pt}\noalign{\smallskip}
\end{tabular}
}
\caption{Validation accuracy on \cub{}. Features learnt from Image-Net class labels transfer well to \cub{}. 
}
\vspace{-0.8cm}
\label{tab:cub-results}
\end{center}
\end{table}

\section{Analysis}
\label{sec:ana}
To better understand the values of visual engagement signals and our pre-trained \vise{} models, we conduct ablation studies and qualitative analysis using the same set of subjective target tasks. All experiments use the fine-tuned protocol unless otherwise specified. 
Additional results and analysis are included in the Appendix~\ref{supsec:result}).
\begin{figure*}[!ht]
\centering
\subfigure[\ube{}.]{
    \includegraphics[scale=0.36]{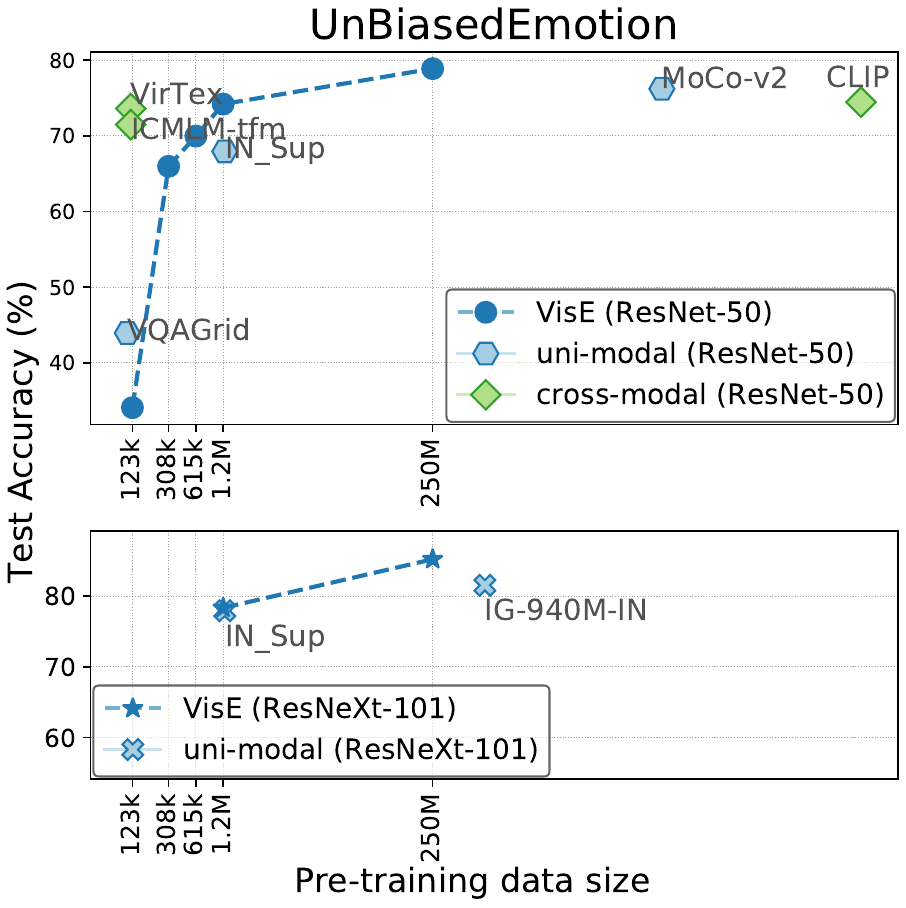}
    \label{fig:size_ube}
}
\hfill
\subfigure[\politics{}.]{
    \includegraphics[scale=0.36]{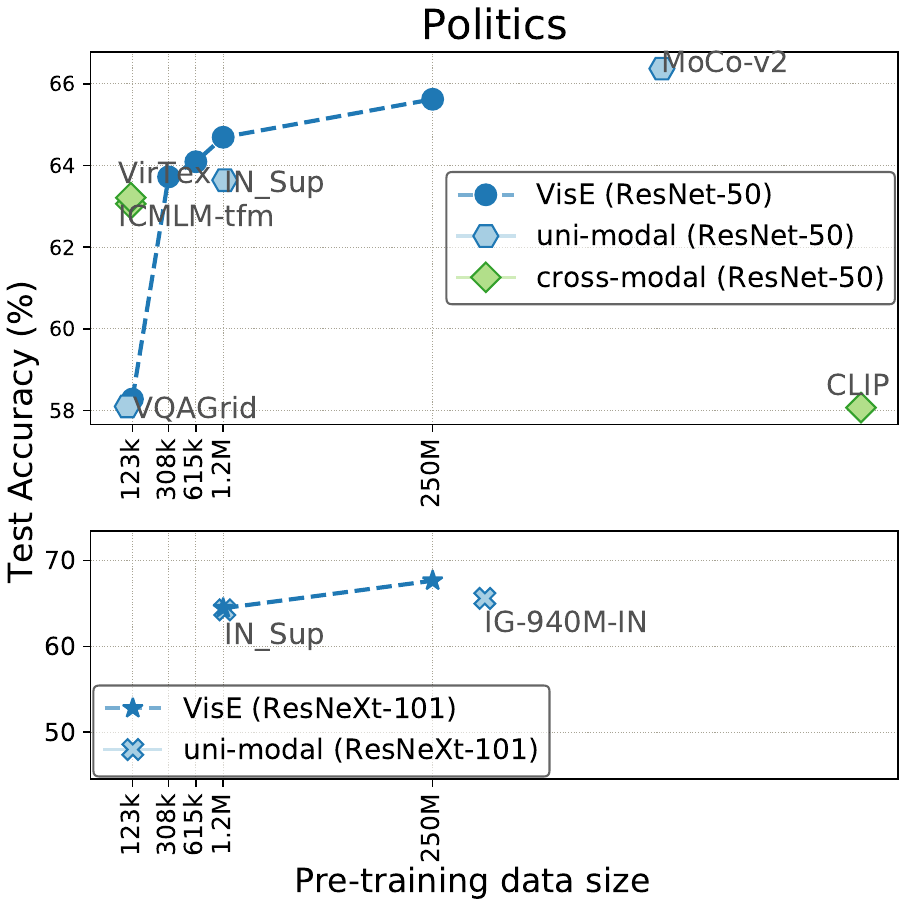}
    \label{fig:size_politics}
}
\hfill
\subfigure[\hm{}.]{
    \includegraphics[scale=0.36]{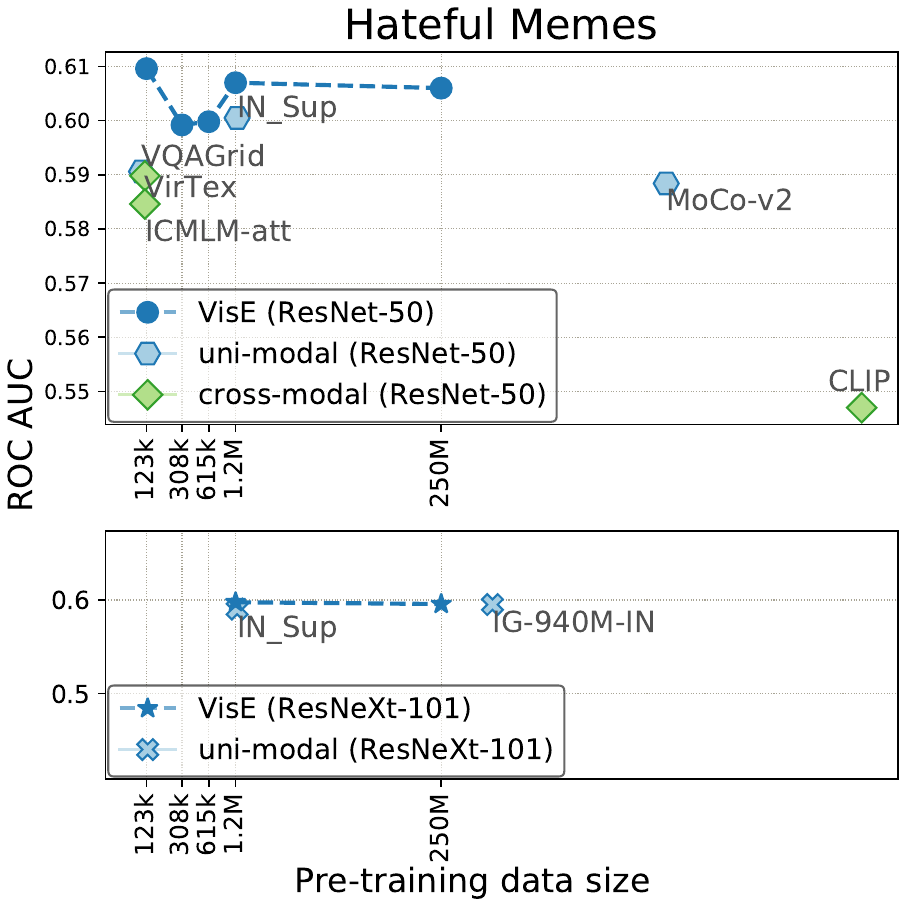}
    \label{fig:size_hm}
}

\caption{Data size ablation using ResNet-50 (\emph{top}) and ResNeXt-101 32$\times$16d (\emph{bottom}) backbones. 
We only present one of \icmlmatt{} and \icmlmtfm{} due to space constraint. 
The differences between both \icmlm{} methods are within 1$\%$ with each other. 
}
\vspace{-0.2cm}
\label{fig:size}
\end{figure*}

\begin{table*}[!htbp]
\scriptsize
\begin{center}
\resizebox{0.9\textwidth}{!}{%
\begin{tabular}{ l l  l l| l l | l l}
\Xhline{0.7pt}\noalign{\smallskip}

\multirow{2}{*}{\textbf{Backbone}} & \multirow{2}{*}{\textbf{Method}} & \multicolumn{2}{ c }{\textbf{\ube{}}} & \multicolumn{2}{ c }{\textbf{\politics{}}}  &\multicolumn{2}{ c }{\textbf{\hm{}}}\\
\cline{3-8}\noalign{\smallskip}
& &\textbf{Linear} & \textbf{Fine-tuned} 
&\textbf{Linear} & \textbf{Fine-tuned}
&\textbf{Linear} & \textbf{Fine-tuned} \\ 
\Xhline{0.7pt}\noalign{\smallskip}

\multirow{3}{*}{ResNet-50} &\reloss{} + \cmloss{}
&\meanstd{45.74}{2.15}   &\meanstd{74.20}{1.93}
&0.6100  &0.6044
&64.30  &64.69
\\
\cline{2-8}\noalign{\smallskip}

&\cmloss{}
&\meanstd{49.15}{1.30}~\Rise{3.41}   &\meanstd{72.16}{1.09}\Drop{2.04}
&0.6005\Drop{0.0095} &0.5921\Drop{-0.0124}
&63.43\Drop{0.87} &63.81\Drop{0.88}
\\

&\reloss{}
&\meanstd{33.05}{1.75} \Drop{12.69}  &\meanstd{70.03}{2.64}\Drop{4.17}
&0.6052\Drop{0.0048}  &0.5980\Drop{0.0064}
&63.09\Drop{1.21}  &63.42\Drop{1.27}
\\





\Xhline{0.7pt}\noalign{\smallskip}
\end{tabular}
}
\caption{
Task ablation of \visesm{} with ResNet-50.
Colored text with \textcolor{green_im}{$\uparrow$} and \textcolor{red}{$\downarrow$} indicate the differences with results from the \vise{} model trained with both reaction and comments.
}
\vspace{-0.3cm}
\label{tab:task_ablation}
\end{center}
\end{table*}
\begin{figure*}[!h]
\centering
\subfigure[\ube{}.]{
    \includegraphics[scale=0.39]{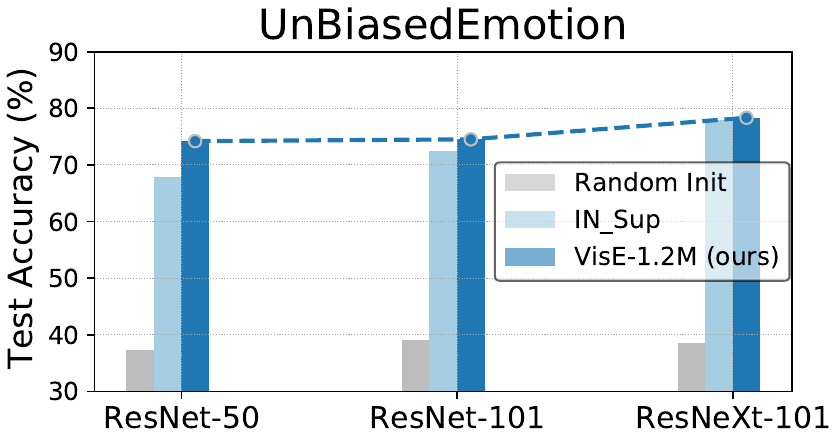}
    \label{fig:backbone_ube}
}
\hfill
\subfigure[\politics{}.]{
    \includegraphics[scale=0.39]{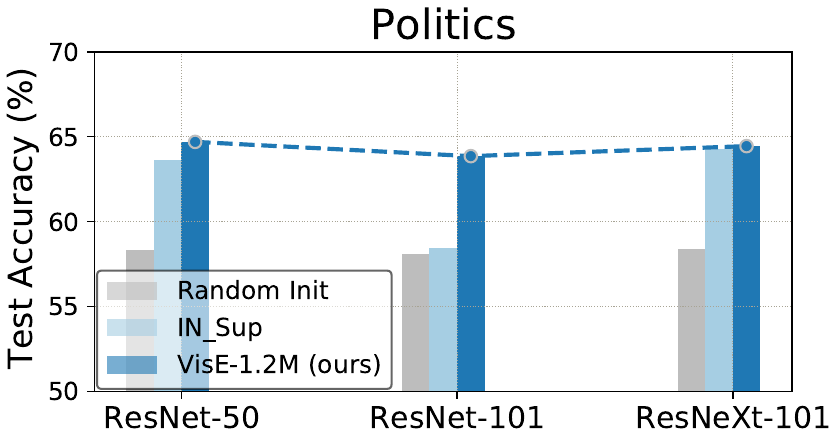}
    \label{fig:backbone_politics}
}
\hfill
\subfigure[\hm{}.]{
    \includegraphics[scale=0.39]{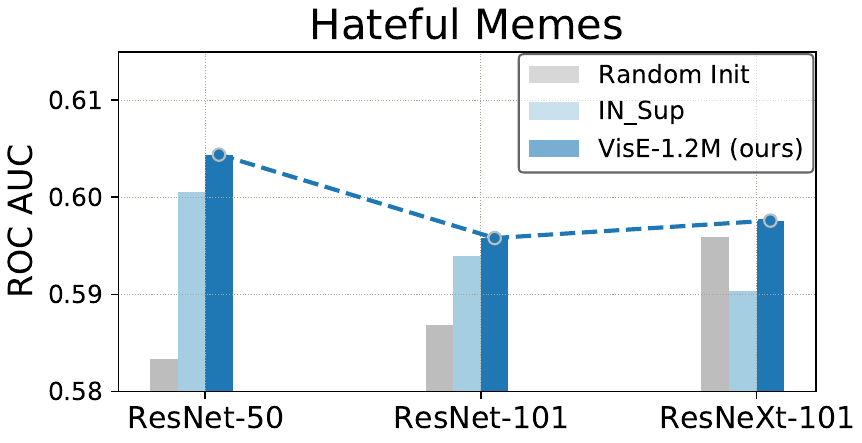}
    \label{fig:backbone_hm}
}
\caption{Visual backbone ablations.
}
\label{fig:backbone}
\end{figure*}
\begin{figure*}[h]
\centering
\subfigure[Image + Text (Fine-tuned)]{
    \includegraphics[scale=0.3]{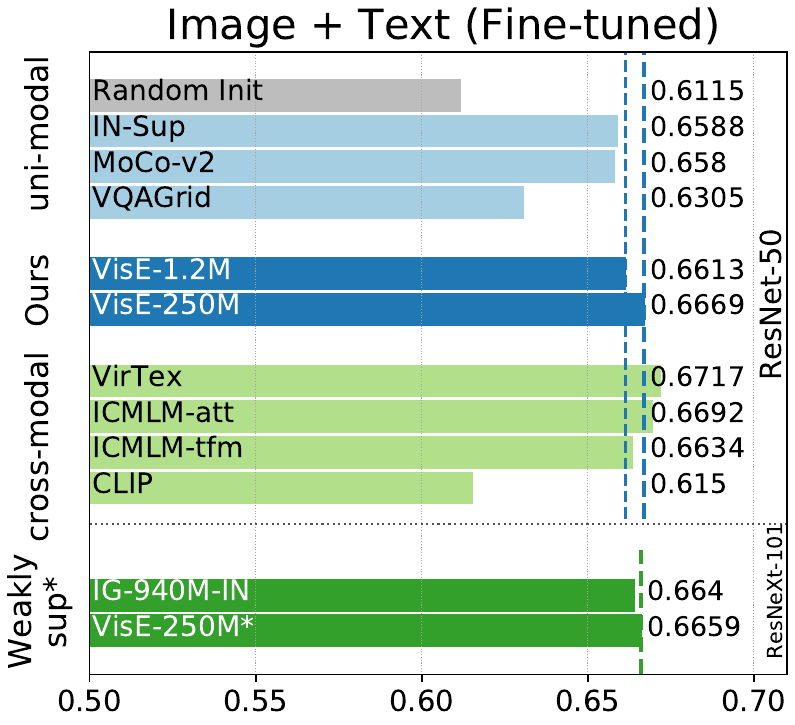}
    \label{fig:mm_full}
}
\hfill
\subfigure[\scriptsize{Image + Text-Frozen (Fine-tuned)}]{
    \includegraphics[scale=0.3]{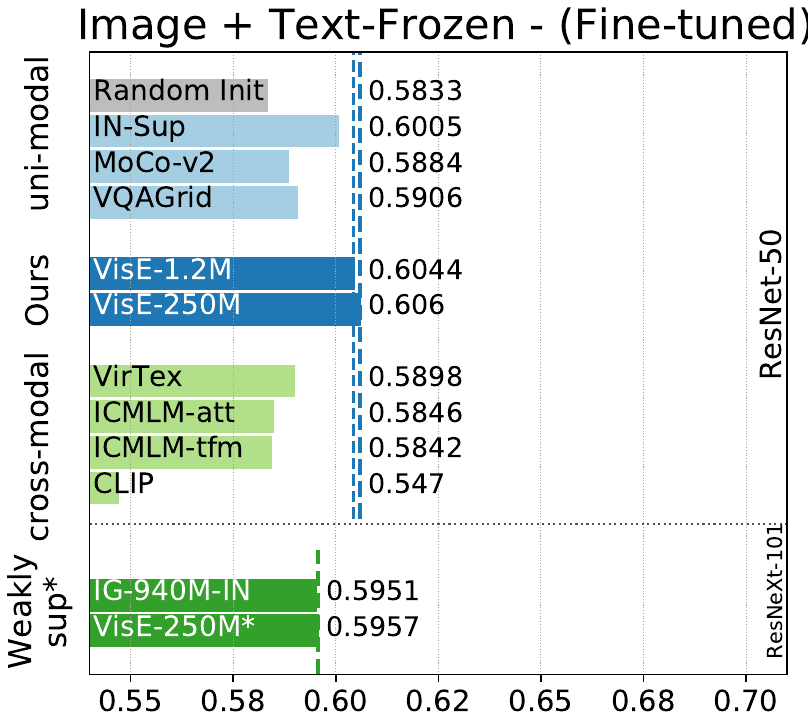}
    \label{fig:mm_full_fztxt}
}
\hfill
\subfigure[Image Only (Fine-tuned)]{
    \includegraphics[scale=0.3]{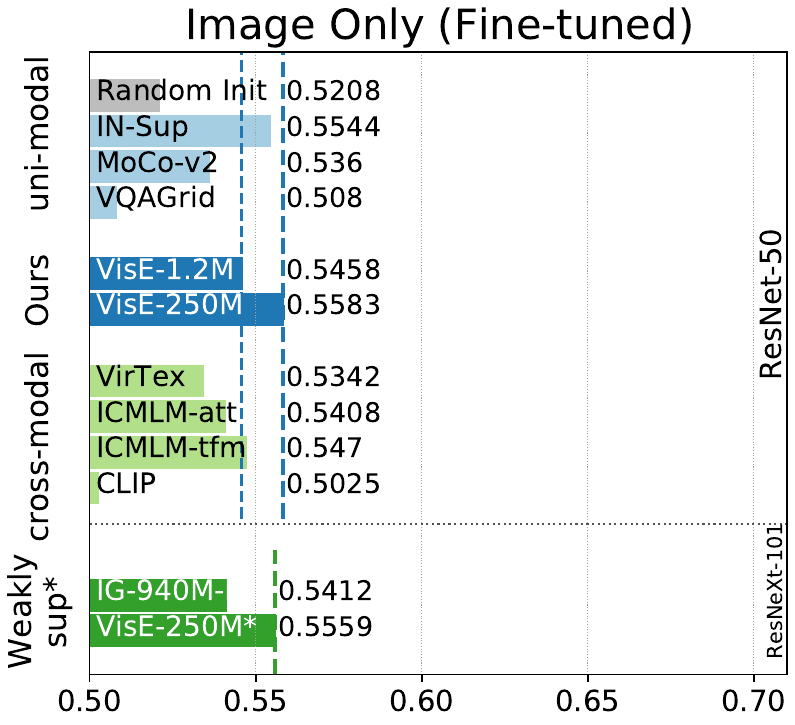}
    \label{fig:mm_img}
}
\hfill
\subfigure[Image Only (Linear)]{
    \includegraphics[scale=0.3]{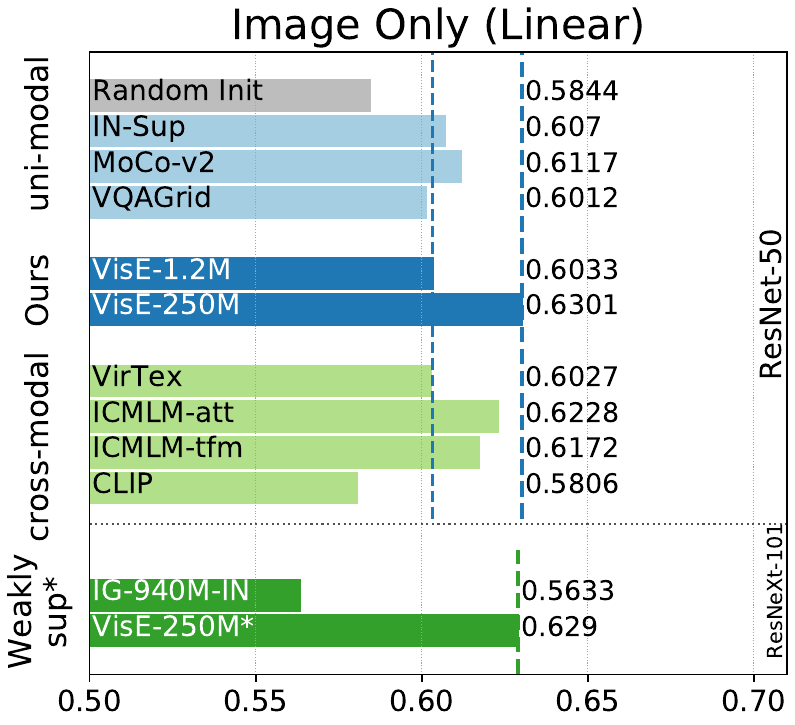}
    \label{fig:mm_imgfz}
}
\caption{Multi-modal ablation for \hm{} \val{} split.
For reference, using text encoder alone give ROC AUC scores: 0.6363 (fine-tuned), 0.5983 (linear).
}
\vspace{-0.2cm}
\label{fig:mm}
\end{figure*}
\begin{figure*}
\centering
\includegraphics[width=0.93\textwidth]{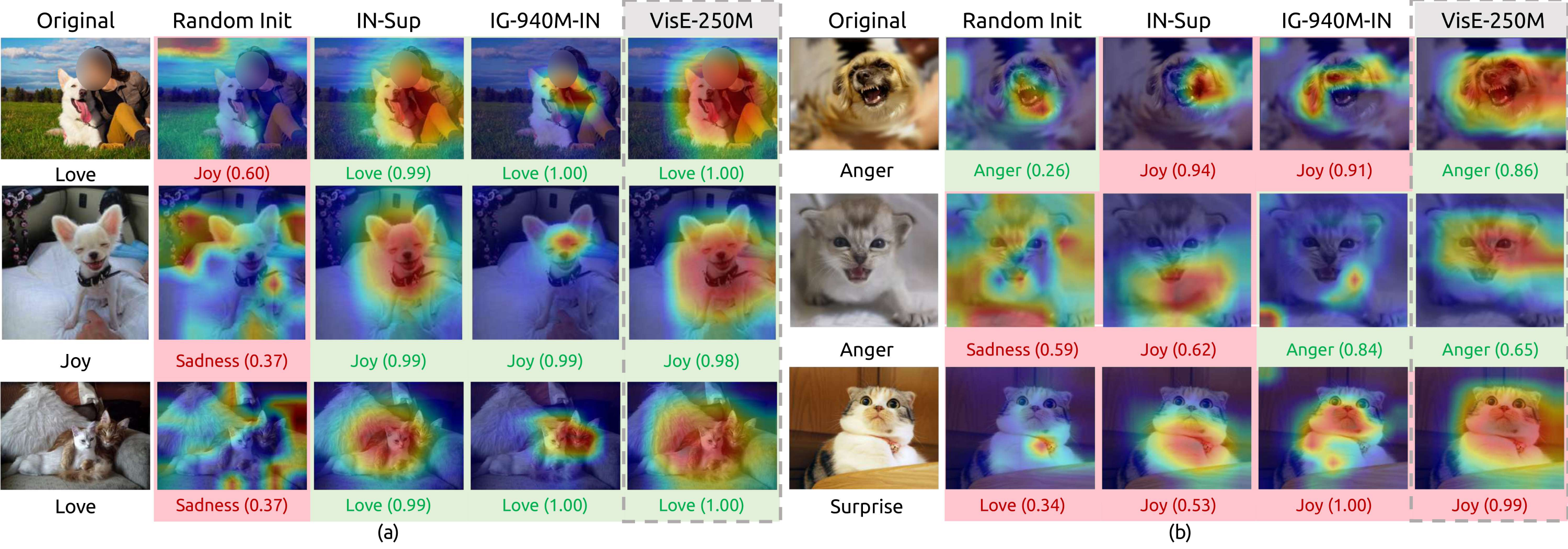}
\caption{
Qualitative results on UnbiasedEmotion dataset using ResNeXt-101 32$\times$16d.
}
\vspace{-0.3cm}
\label{fig:cam}
\end{figure*}

\subsection{Pre-training ablation}\label{subsec:ana_data}
\paragraph{Effect of pre-training data size}
We additionally train \vise{} models using randomly sampled $\{$123k, 308k, 615k$\}$ images using fine-tuned setting.
Fig.~\ref{fig:size} presents the results for three datasets compared with other baselines for easy reference.
From Fig.~\ref{fig:size_ube} and~\ref{fig:size_politics}, we can see that there is a positive correlation between training data size and model performance for \vise{}. 
\virtex{} and \icmlm{} achieve similar results as \vise{} with less training data (118k \vs $\{$615k, 1.23M$\}$). 
This shows the noisy pseudo-labels from visual engagement might need more data to compensate its weakly-supervised nature.

For a multi-modal dataset like \hm{}, the correlation between data size and performance is less clear than the other two tasks, as shown in Fig.~\ref{fig:size_hm}. When fine-tuned, \vise{} is able to perform better than other baselines, demonstrating the effectiveness of visual engagement signals.

\paragraph{Effect of pre-training tasks}
\vise{} is trained using cluster assignments from both comments and raw reactions.
Table~\ref{tab:task_ablation} shows the ablation study where we evaluate the contribution of different engagement formats.
In general, \vise{} obtains the best performance when trained with a multi-task objective.

\cvpara{Effect of pre-training visual backbone}\qquad
Fig.~\ref{fig:backbone} presents ablation studies using different visual backbones.
\visesm{} is better than \random{} and \insup{} across all three backbone choices and three downstream tasks. We also note that the advantages of \vise{} is diminishing as the number of parameters of backbone getting larger possibly due to overfitting on the downstream training set.

\subsection{Multimodal fine-tuning ablation}\label{subsec:ana_mulmodal}
We used an image encoder and a frozen textual encoder for experiments on \hm{} dataset in Sec~\ref{sec:exp}. To isolate the effects of both module for this dataset, we compare \vise{} with other baselines under the following setups:
(1) Image + Text (Fine-tuned): parameters from both encoders are updated during transfer learning.
(2) Image + Text-Frozen (Fine-tuned): the same fine-tuned setting as the experiment in Sec~\ref{sec:exp}.
(3) Image Only (Fine-tuned): we only use image encoder in fine-tuned setting.
(4) Image Only (Linear): image encoder is used as feature extractor only.
Note the linear evaluation of \hm{} in Sec.~\ref{subsec:expmain} (Fig.~\ref{fig:linear_hm}) uses concatenated features from both encoders.
Fig.~\ref{fig:mm} presents the results. 

Methods with \vise{} are able to achieve better or comparable results compared with other baselines. And visual-language methods outperform \insup{} in Fig.~\ref{fig:mm_full} except for \clip{}. This shows the visual backbones that are pre-trained with a textual module are better when both modules are fine-tuned together.

From Figs.~\ref{fig:mm_full}-\ref{fig:mm_img}, the ROC AUC scores are getting smaller and smaller as the textual encoder makes less and less contribution. 
This demonstrates that the text module has dominant effect on predicting if a multi-modal meme is hateful or not. If using image encoders only, linear evaluation obtains better results than the fine-tuned protocol (Fig.~\ref{fig:mm_img} \vs~\ref{fig:mm_imgfz}). This also suggests textual information is more important on this dataset.

\subsection{Qualitative Analysis}\label{subsec:ana_qual}
We also conduct qualitative studies using \ube{} to further understand the benefit of visual engagement. 
Fig.~\ref{fig:cam} shows sample predictions produced by our \viselg{} and 3 other approaches. 
The predicted classes for each approach are color coded (\textcolor{green_im}{green} as correct, \textcolor{red}{red} as incorrect).
We also use class activation mappings~\cite{zhou2015cnnlocalization} to visualize the discriminate image regions for the predicted emotion.
Although all methods that are initialized with pre-trained models can detect the object of interest in the image,
\insup{} and \wsl{} are more likely to predict ``joy'' and ``love'' for the cats and dogs photos in Fig.~\ref{fig:cam}, while \viselg{} yields more diverse predictions (see row 2 left \vs row1 right as an example). 
This seems to suggest that \insup{} and \wsl{} map dogs and cat to positive emotions. 
\viselg{}, on the other hand, does not rely on the ImageNet object labels during pre-training and is able to distinguish the subtle emotional differences among different images with dogs and cats.
However, all methods failed to infer ``surprise'' from the bottom left image, possibly due to imbalanced training data of \ube{}.
Additional visualization can be found in the Appendix~\ref{supsec:vis}.

\section{Conclusion}
We explored social media visual engagement as supervisory signals for representation learning. We presented \vise{}, a streamlined pre-training method that uses pseudo-labels derived from human responses to social media posts, including reactions and comments. Experiments and analysis show that visual engagement signals transfer well to various downstream tasks that go beyond conventional visual recognition. \vise{} is able to outperform various representation learning models on these datasets.  
We therefore hope that \vise{} could inspire and facilitate future research that focuses on the cognitive aspects of images.
Pre-trained models will be released upon acceptance of the work.

\small{
\cvpara{Acknowledgement}\quad
We thank Marseille 
who is featured in Figs.~\ref{fig:teaser} and~\ref{fig:method}.
This work is supported by a Facebook AI research grant awarded to Cornell University.
}

\appendix

\section{Qualitative Analysis}
\label{supsec:vis}

\begin{figure}[t]
\centering
\includegraphics[width=\columnwidth]{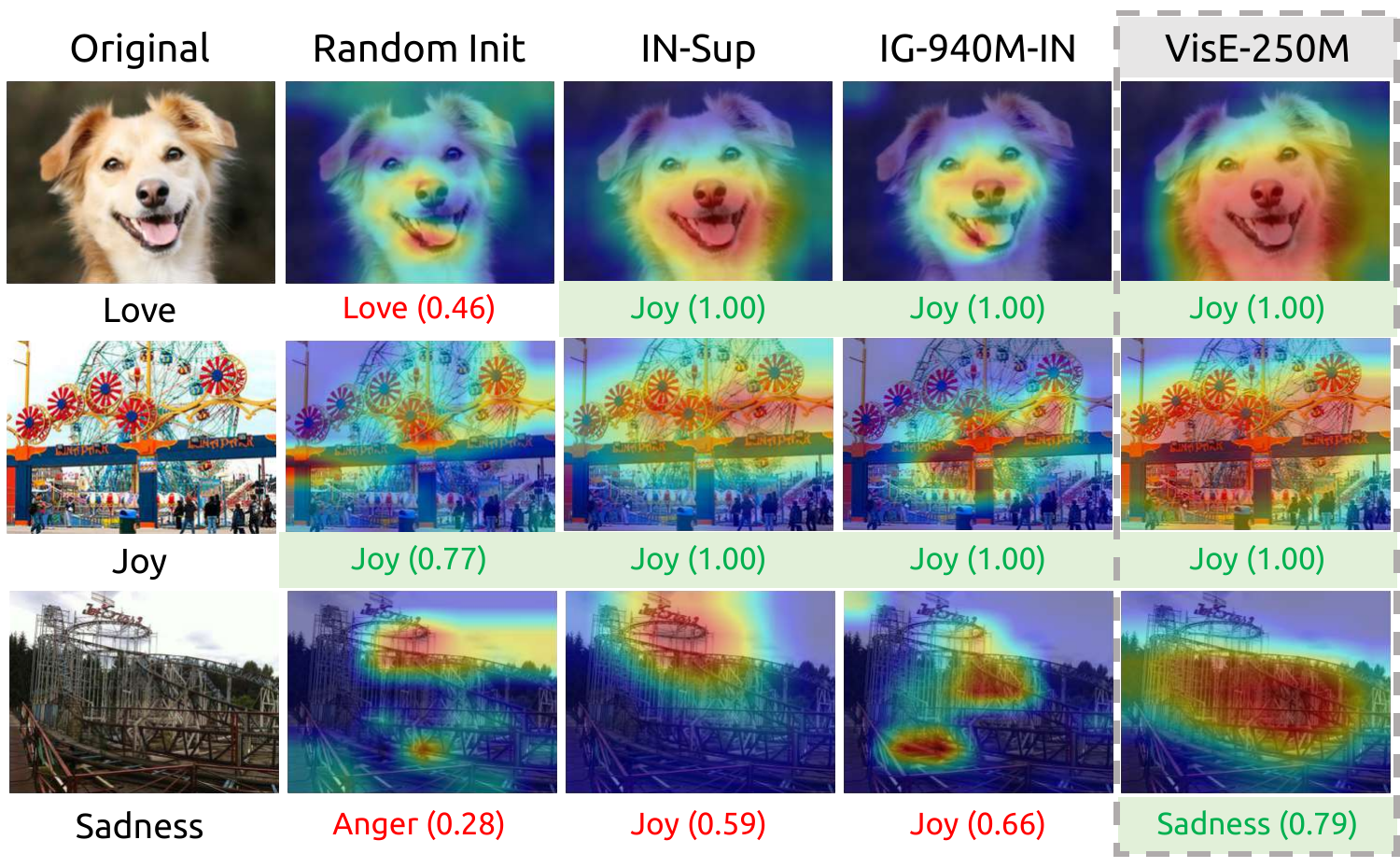}
\caption{
Qualitative results on \ube{} dataset using ResNeXt-101 32$\times$16d backbone.
}
\label{supfig:cam}
\end{figure}

Fig.~\ref{supfig:cam} presents more predicted examples including images with dogs and parks. It further shows that ImageNet based pre-training methods tend to map certain objects to a certain set of emotions. \vise{}, on the other hand, is able to predict correct emotions in these examples.

\section{Supplementary Results and Discussion}
\label{supsec:result}

\paragraph{Transfer learning on ImageNet} Table~\ref{supptab:imagenet} presents results and comparisons on ImageNet. We fine-tune \viselg{} with a ResNeXt-101 backbone on ImageNet, and compare the \val{} accuracy scores with the same ResNeXt-101 model trained from scratch (\insup{}). We also show the results of \wsl{}~\cite{wslimageseccv2018}, which is pre-trained on 940 million images with 1.5K hashtags and fine-tuned on ImageNet using the same visual backbone.
We see from Table~\ref{supptab:imagenet} that representations learned from \viselg{} with engagement signals are transferable to ImageNet, outperforming the \insup{} model by 0.88 (1.12\%) measured by Top-1 accuracy. Note that engagement signals are relatively weak compared to the hashtags used in \wsl{}, which were selected to match with 1000 ImageNet synsets. Our goal here is to show features learned by \vise can be generalized to large-scale image classification tasks.

\paragraph{Images \vs engagement signals} 
To disentangle the effect of training images and engagement signals, we also trained \moco{} with the same 1.23 million social post data (\visesm{}(\moco{})). Table~\ref{supptab:ssl} shows the linear evaluation results on \ube{}, which
shows the engagement signals, not the images, are beneficial for this dataset. We will include the full results in the final version.

\paragraph{Additional results} 
Table~\ref{supptab:linear_results} and~\ref{supptab:ft_results} present full transfer learning results including performance on the \val{} split and an additional metric for the \hm{} dataset. 
These two tables can be read in conjunction with the main figure and the backbone ablation studies in the main text.
Note that we use in-house baselines instead of copying results from prior work for fair-comparison purposes.
All the experiments are trained using the same grid search range, validation set, learning rate schedule, \etc.
We use validation accuracy and ROC AUC for \hm{} to select the best set of hyper-parameters. See Appendix~\ref{supsubsec:ft_details} for details.

\paragraph{Datasize calculation for contrastive learning methods}
In size ablation studies, we sort all pre-training methods by the training inputs size. 
We consider the negative input pairs for \moco{} and \clip{} as the \emph{effective} training data size.

\begin{itemize}
    \item \textbf{\moco{}} uses image pairs from ImageNet as inputs. The total class size is the total number of training data (1.28 million). The effective training data size is the number of image pairs used, which is (k + 1) $\times$ 1.28M = 83.9B, where $k=65536$ is the number of images in the queue for \moco{}.
    \item \textbf{\clip{}} uses a dataset with 400M image-text pairs. This approach considers the pair-wise similarity among image-text in a batch during training.
Since the batch size is 32768, the total effective datasize is $400$M $/ 32768 \times 32768 \times 32768=83.9$B.
\end{itemize}

\begin{table}
\small
\begin{center}
\begin{tabular}{ l l  l }
\Xhline{0.7pt}\noalign{\smallskip}
\textbf{Method} &\textbf{Top 1 Accuracy} &\textbf{Top 5 Accuracy}
\\
\Xhline{0.7pt}\noalign{\smallskip}
\insup{} &78.78	& 94.12 \\
\viselg{} &79.66\Rise{0.88}	&94.62 \Rise{0.5} \\
\hline

\wsl{}~\cite{wslimageseccv2018} &84.2	&97.2 \\

\Xhline{0.7pt}\noalign{\smallskip}
\end{tabular}
\caption{
Fine-tuned experiments on ImageNet with ResNeXt-101 32 $\times$ 16d backbone.
Colored text with \textcolor{green_im}{$\uparrow$} indicate the differences between \vise{} and \insup{}.
}
\label{supptab:imagenet}
\vspace{-0.3cm}
\end{center}
\end{table}

\begin{table}
\small
\begin{center}
\resizebox{\columnwidth}{!}{%
\begin{tabular}{ l l  l }
\Xhline{0.7pt}\noalign{\smallskip}
\textbf{Method} &\textbf{Val Accuracy} &\textbf{Test Accuracy}
\\
\Xhline{0.7pt}\noalign{\smallskip}
\visesm{} (\moco{}) &\meanstd{27.96}{2.91} & \meanstd{27.80}{2.30} \\
\visesm{} &\meanstd{44.67}{3.52} \Rise{16.71}	&\meanstd{45.74}{2.15} \Rise{17.94} \\
\Xhline{0.7pt}\noalign{\smallskip}
\end{tabular}
}
\caption{
Linear evaluation experiments on \ube{} with ResNet-50 using the same 1.23 million data.
}
\label{supptab:ssl}
\vspace{-0.3cm}
\end{center}
\end{table}

\begin{table*}
\small
\begin{center}
\resizebox{0.9\textwidth}{!}{%
\begin{tabular}{@{\extracolsep{4pt}} l l  l l l l l l@{}}
\Xhline{0.7pt}\noalign{\smallskip}

\multirow{2}{*}{\textbf{Backbone}} & \multirow{2}{*}{\textbf{Method}} & \multicolumn{2}{ c }{\textbf{\ube{}}} & \multicolumn{2}{ c }{\textbf{\politics{}}}  &\multicolumn{2}{ c }{\textbf{\hm{}}}\\
\cline{3-4}
\cline{5-6}
\cline{7-8}\noalign{\smallskip}
& &\textbf{Val Accuracy} & \textbf{Test Accuracy} &\textbf{Val Accuracy} & \textbf{Test Accuracy}  &\textbf{ROC AUC} &\textbf{Accuracy} \\ 
\Xhline{0.7pt}\noalign{\smallskip}

\multirow{14}{*}{ResNet-50} &Random Init
&\meanstd{24.67}{2.78}  &\meanstd{23.80}{1.02}
&56.80	&56.57
&0.5335 &51.64	
\\

&\multicolumn{7}{ c }{\graycell Uni-modality pre-training methods} \\
&\insup{}
&\meanstd{43.62}{2.31}	&\meanstd{44.36}{1.09}
&59.31	&59.45
&0.5691 &53.16	
\\

&\grid{}~\cite{jiang2020defense}
&\meanstd{32.17}{1.22}	&\meanstd{33.57}{0.86}
&57.32	&57.31
&0.5517  &53.2\Rise{0.04}
\\

&\multicolumn{7}{ c }{\graycell Cross-modalities pre-training methods}
\\
&\virtex{}~\cite{desai2021virtex} 
&\meanstd{40.59}{2.96}	&\meanstd{42.17}{1.14}
&58.46	&58.44
&0.5659 &54.40\Rise{1.24}
\\
&\icmlmatt{}~\cite{bulent2020icmlm} 
&\meanstd{23.81}{2.25}	&\meanstd{23.51}{1.52}
&58.27	&58.41
&0.5702\Rise{0.0011} &53.32\Rise{0.16}	
\\
&\icmlmtfm{}~\cite{bulent2020icmlm} 
&\meanstd{31.71}{2.02}	&\meanstd{31.87}{0.95}
&58.73	&58.86
&0.5631 &53.24\Rise{0.08}
\\
&\multicolumn{7}{ c }{\graycell Contrastive learning pre-training methods}
\\
&\moco{}~\cite{chen2020mocov2}
&\meanstd{26.31}{1.12}	&\meanstd{26.23}{1.20}
&58.14	&58.30
&0.5947\Rise{0.0256} &53.92	\Rise{0.76}	
\\
&\clip{}~\cite{radford2021learning}
&\meanstd{42.70}{3.02}	&\meanstd{45.41}{2.90}\Rise{1.05}
&56.65	&56.42
&\textbf{0.6147}\Rise{0.0456} &\textbf{57.04}\Rise{3.88}	
\\
&\multicolumn{7}{ c }{\graycell Ours}
\\

&\visesm{}
&\meanstd{44.67}{3.52}\Rise{1.05}	&\meanstd{45.74}{2.15}\Rise{1.38}
&59.15\Drop{0.16}	&59.30\Drop{0.15}
&0.6100\Rise{0.0409}	 &55.52\Rise{2.36}	
\\	
&\viselg{}
&\meanstd{\textbf{51.97}}{4.08}\Rise{8.35}	&\meanstd{\textbf{53.05}}{1.48}\Rise{8.69}
&\textbf{60.56}\Rise{1.25}	&\textbf{60.31}\Rise{0.86}	
&0.5784\Rise{0.0093} &54.48\Rise{1.32}	
\\
\Xhline{0.7pt}\noalign{\smallskip}

\multirow{7}{*}{\shortstack[l]{ResNeXt-101\\$32 \times 16$d}} &Random Init  
&\meanstd{37.96}{3.77}	&\meanstd{38.43}{1.38}
&57.05	&56.92
&0.5466 &53.64
\\
\cline{2-8}\noalign{\smallskip}
&\insup{}
&\meanstd{63.09}{3.12}	&\meanstd{62.59}{1.99}
&59.24	&59.42
&0.5542 &51.84	
\\
&\wsl{}~\cite{wslimageseccv2018} 
&\meanstd{55.86}{1.36}	&\meanstd{56.26}{1.32}
&60.98\Rise{1.74}	&\textbf{61.15}\Rise{1.73}
&0.5482 &52.28\Rise{0.44}
\\

&\visesm{} (ours)
&\meanstd{56.64}{2.49}\Drop{6.45}	&\meanstd{56.26}{1.05}\Drop{6.33}
&59.70\Rise{0.46}	&59.89\Rise{0.47}
&0.5621\Rise{0.0079} &54.24\Rise{2.40}
\\
&\viselg{} (ours)
&\meanstd{\textbf{69.61}}{2.74}\Rise{6.51}	&\meanstd{\textbf{69.44}}{1.20}\Rise{6.85}
&\textbf{61.08}\Rise{1.84}	&61.01\Rise{1.59}
&\textbf{0.5795}\Rise{0.0253} &\textbf{56.04}\Rise{4.20}	
\\
\Xhline{0.7pt}\noalign{\smallskip}
\end{tabular}
}
\caption{
Linear evaluation experiments comparing \vise{} with other pre-training baselines.
Colored text with \textcolor{green_im}{$\uparrow$} and \textcolor{red}{$\downarrow$} indicate the differences between \vise{} and \insup{} with the same visual backbone.
\textcolor{green_im}{$\uparrow$} is also used if other methods yield better results than \insup{}. 
In general, \vise{} outperforms the ImageNet supervised and hashtag-based weakly supervised pre-training methods.
}
\label{supptab:linear_results}
\end{center}
\end{table*}

\begin{table*}
\small
\begin{center}
\resizebox{0.9\textwidth}{!}{%
\begin{tabular}{@{\extracolsep{4pt}} l l  l l l l l l@{}}
\Xhline{0.7pt}\noalign{\smallskip}

\multirow{2}{*}{\textbf{Backbone}} & \multirow{2}{*}{\textbf{Method}} & \multicolumn{2}{ c }{\textbf{\ube{}}} & \multicolumn{2}{ c }{\textbf{\politics{}}}  &\multicolumn{2}{ c }{\textbf{\hm{}}}\\
\cline{3-4}
\cline{5-6}
\cline{7-8}\noalign{\smallskip}
& &\textbf{Val Accuracy} & \textbf{Test Accuracy} &\textbf{Val Accuracy} & \textbf{Test Accuracy}  &\textbf{ROC AUC} &\textbf{Accuracy} \\ 
\Xhline{0.7pt}\noalign{\smallskip}

\multirow{14}{*}{ResNet-50} &Random Init  
&\meanstd{39.01}{0.99}	&\meanstd{37.25}{2.12}	
&58.28	&58.31 
&0.5833 &51.84	\\

&\multicolumn{7}{ c }{\graycell Uni-modality pre-training methods} \\
&\insup{}
&\meanstd{69.87}{3.27}	&\meanstd{67.94}{3.18}	
&63.87	&63.64 
&0.6005       &54.32
\\
&\grid{}~\cite{jiang2020defense}
&\meanstd{42.17}{3.07}	&\meanstd{43.93}{1.56}	
&58.31	&58.1
&0.5906 &53.24
\\

&\multicolumn{7}{ c }{\graycell Cross-modalities pre-training methods} \\

&\virtex{}~\cite{desai2021virtex} 
&\meanstd{72.24}{2.13}\Rise{2.37}	&\meanstd{73.61}{1.94}\Rise{5.67}
&63.24	&63.06
&0.5898	&53.84	
\\

&\icmlmatt{}~\cite{bulent2020icmlm} 
&\meanstd{71.65}{2.31}\Rise{1.78}	&\meanstd{70.98}{2.01}\Rise{3.05}
&63.3	&63.2
&0.5846	&53.52
\\

&\icmlmtfm{}~\cite{bulent2020icmlm} 
&\meanstd{70.92}{1.65}\Rise{1.05}	&\meanstd{71.48}{1.78}\Rise{3.54}
&63.43	&63.21
&0.5842	 &53.40
\\
&\multicolumn{7}{ c }{\graycell Contrastive learning pre-training methods}
\\
&\moco{}~\cite{chen2020mocov2}
&\meanstd{77.63}{1.78}\Rise{7.76}	&\meanstd{76.23}{1.88}\Rise{8.29}	
&\textbf{66.24}\Rise{2.37}	&\textbf{66.37}\Rise{2.73}
&0.5884   &52.48	
\\
&\clip{}~\cite{radford2021learning}
&\meanstd{73.68}{0.93}\Rise{3.81}	&\meanstd{74.46}{1.21}\Rise{6.53}
&58.08	&58.07
&0.5470  &53.48	
\\
&\multicolumn{7}{ c }{\graycell Ours} \\

&\visesm{}
&\meanstd{73.82}{1.07}\Rise{3.95}	&\meanstd{74.20}{1.93}\Rise{6.26}
&64.69\Rise{0.82}        	&64.69\Rise{1.05}
&\textbf{0.6070}\Rise{0.0039}	 &\textbf{55.88}\Rise{0.96}	\\
	
&\viselg{}
&\meanstd{\textbf{79.74}}{1.54}\Rise{9.87}	&\meanstd{\textbf{78.89}}{2.23}\Rise{10.95}
&65.83\Rise{1.96}	&65.62\Rise{1.98}
&0.6060\Rise{0.0055}	&55.00\Rise{0.68}
\\
\Xhline{0.7pt}\noalign{\smallskip}

\multirow{3}{*}{ResNet-101} &Random Init  
&\meanstd{40.20}{2.76}	&\meanstd{39.08}{1.72}	
&58.18	&58.07 
&0.5868  &53.48	\\

&\insup{}
&\meanstd{71.84}{2.72}	&\meanstd{72.43}{2.24}	
&58.28	&58.42 
&0.5939  &\textbf{54}  \\

&\visesm{} (ours)
&\meanstd{\textbf{73.82}}{0.77}\Rise{1.97}	&\meanstd{\textbf{74.52}}{1.18}\Rise{2.10}	
&\textbf{63.92}\Rise{5.64}	&\textbf{63.85}\Rise{5.43}
&\textbf{0.5958}\Rise{0.0019}	  &52.96\Drop{1.04}	
 \\

\Xhline{0.7pt}\noalign{\smallskip}

\multirow{7}{*}{\shortstack[l]{ResNeXt-101\\$32 \times 16$d}} &Random Init  
&\meanstd{40.20}{2.65}	&\meanstd{38.59}{0.91}	
&58.26	&58.39 
&0.5959  &\textbf{54.68} \\
\cline{2-8}\noalign{\smallskip}

&\insup{}
&\meanstd{79.00}{2.33}	&\meanstd{77.92}{2.38}
&64.22	&64.25 
&0.5903  &52.92	
\\
&\wsl{}~\cite{wslimageseccv2018} 
&\meanstd{83.24}{1.68}\Rise{4.24}	&\meanstd{81.52}{1.76}\Rise{3.60}
&65.90\Rise{1.68}	&65.58\Rise{1.33}
&0.5951\Rise{0.0048}  &54.28\Rise{1.36}		
\\

&\visesm{} (ours)
&\meanstd{77.57}{2.43}\Drop{1.43}	&\meanstd{78.33}{1.39}\Rise{0.41}
&64.61\Rise{0.39}	   &64.44\Rise{0.19}
&0.5976\Rise{0.0073}  &54.40\Rise{1.48}	
\\
&\viselg{} (ours)
&\meanstd{\textbf{84.08}}{1.87}\Rise{5.08}	&\meanstd{\textbf{85.21}}{1.24}\Rise{7.29}
&\textbf{67.61}\Rise{3.39}	   &\textbf{67.64}\Rise{3.39}
&\textbf{0.5957}\Rise{0.0054}  &\textbf{54.96}\Rise{2.04}	\\

\Xhline{0.7pt}\noalign{\smallskip}
\end{tabular}
}
\caption{
Fine-tuning experiment comparing \vise{} with other pre-training baselines. 
Colored text with \textcolor{green_im}{$\uparrow$} and \textcolor{red}{$\downarrow$} indicate the differences with \insup{} with the same visual backbone. 
\textcolor{green_im}{$\uparrow$} is also used if other methods yield better results than \insup{}. 
Similar to observations in Table~\ref{supptab:linear_results}, \vise{} can achieve better results compared to the ImageNet supervised and hashtag-based weakly supervised pre-training methods.
}
\label{supptab:ft_results}
\end{center}
\end{table*}

\section{Reproducibility Details}
\label{supsec:detail}

\subsection{\vise{} Pre-training Setup}

\paragraph{Optimization} \vise{} models are trained on 32 GPUs across 4 machines with a batch size of 1920 images for the ResNet backbone and 1536 for the ResNeXt backbone.
We use stochastic gradient descent with a momentum of 0.9 and a weight decay of 0.0001.
The base learning rate is set according to $0.1 / 256 \times b$, where $b$ is the batch size used for the particular model. 
The learning rate is warmed up linearly from 0 to the base learning rate during the first $5\%$ of the whole iterations.
The learning rate decay schedule is set differently for \visesm{} and \viselg{}. For models that use 1.23 million images, we follow common ImageNet pre-training settings.
For models that are trained with 250 million images, 
the learning rate is reduced 10 times over approximately 10 epochs with the scaling factor of 0.5.

\paragraph{Training details}
We adopt standard image augmentation strategy during training (randomly resize crop to 224 $\times$ 224 and random horizontal flip).
Since the dataset is not balanced, we follow~\cite{focal_loss,cui2019cbloss} to stabilize the training processing by initializing the the bias for the last linear classification layer  with $b = -\log\left(\left(1 - \pi\right)/\pi\right)$, where the prior probability $\pi$ is set to 0.01.
To obtain the pseudo-labels for the visual engagement signals, we set the number of clusters as 5000 and 128, for comments and raw reactions respectively.

\paragraph{Other details}
We spend around 9 hours to mine the data for pretraining with 3 server nodes (144 cpus).
For the 1.23M data, the total word count for comments is 178M, the \meanstd{average}{std} number of comments per image is \meanstd{20.25}{54.43}, the \meanstd{average}{std} reactions count per image is \meanstd{81.21}{601.4}. 
We use Pytorch~\cite{paszke2017pytorch} to implement and train all the models on NVIDIA Tesla V100 GPUs.

\begin{table*}[h]
\small
\begin{center}
\resizebox{0.9\textwidth}{!}{%
\begin{tabular}{l l  l l l l}
\Xhline{0.7pt}\noalign{\smallskip}
\textbf{Dataset}   &\textbf{Task}  & \textbf{\# Classes}    &\textbf{Train}  &\textbf{Val}  &\textbf{Test}\\ 
\Xhline{0.7pt}\noalign{\smallskip}

\longcub{}~\cite{WahCUB_200_2011}
& Fine-grained bird species recognition
&200
&5994	&5794	&- \\
\hline\noalign{\smallskip}

\ube{}~\cite{panda2018ube}
& Image emotion recognition
&6
&2,131$^{\star}$	&304$^{\star}$	&610$^{\star}$ \\
\hline\noalign{\smallskip}

\politics{}~\cite{NEURIPS2019_politics}
 &Visual political
bias prediction  &2 &  607,306$^{\star}$ & 67,478$^{\star}$ & 75,148 \\
\hline\noalign{\smallskip}

\hm{}~\cite{kiela2020hateful}
& Hate speech detection in multimodal memes  &2  & 8,500	& 500  & - \\
\Xhline{0.7pt}\noalign{\smallskip}
\end{tabular}
}
\caption{Specifications of the various target task dataset. Image number with $^{\star}$ are the subset we randomly sampled since no publicly data splits are available.
\ube{} are randomly split 5 times.
}
\vspace{-0.4cm}
\label{tab:datasets}
\end{center}
\end{table*}

\begin{table*}
\small
\begin{center}
\resizebox{0.9\textwidth}{!}{%
\begin{tabular}{@{\extracolsep{4pt}} l  l l l l l l l l @{}}
\Xhline{0.7pt}\noalign{\smallskip}
\multirow{3}{*}{\textbf{Task}}
&\multirow{3}{*}{\textbf{\# GPUs}}
& \multicolumn{2}{ c }{\textbf{ResNet-50 (24M)}}
& \multicolumn{2}{ c }{\textbf{ResNet-101 (45M)}}
& \multicolumn{2}{ c }{\textbf{ResNeXt-101 (194M)}}

\\
\cline{3-4}
\cline{5-6}
\cline{7-8}
& &\multirow{2}{*}{\textbf{\shortstack[l]{Per Iteration \\ (second)}}} &\multirow{2}{*}{\textbf{\shortstack[l]{Total Time \\(minute)}}}
&\multirow{2}{*}{\textbf{\shortstack[l]{Per Iteration \\ (second)}}} &\multirow{2}{*}{\textbf{\shortstack[l]{Total Time \\(minute)}}}
&\multirow{2}{*}{\textbf{\shortstack[l]{Per Iteration \\ (second)}}} &\multirow{2}{*}{\textbf{\shortstack[l]{Total Time \\(minute)}}}
\\ 
& & & & & & & \\
\Xhline{0.7pt}\noalign{\smallskip}
\ube{} &1
&1.67	&49.64
&1.44 	&50.30
& 1.32	&63.37
\\
\politics{} &\multirow{1}{*}{8} &0.14	&216.26 
&0.18  &243.64 
&0.67	&721.61 
\\

\hm{} &1 &0.43	&41.23
&0.32	&49.57 
&0.50	&140.81 
\\

\cub{} &1 &0.74 &294.01 &- &-
&0.91 &1,304.01  
\\

\Xhline{0.7pt}\noalign{\smallskip}
\end{tabular}
}
\caption{
Average run time (per iteration and total) for fine-tuned experiments.
}
\vspace{-0.4cm}
\label{supptab:infra}
\end{center}
\end{table*}

\begin{table*}[!htbp]
\scriptsize
\begin{center}
\resizebox{\textwidth}{!}{%
\begin{tabular}
{@{\extracolsep{4pt}} l l l ll l l l l l l l l l@{}}
\Xhline{0.7pt}\noalign{\smallskip}
& \multirow{2}{*}{\textbf{\shortstack{Training\\Schedule}}}
& \multirow{2}{*}{\textbf{}} 
& \multirow{2}{*}{\textbf{Method}}
& \multirow{2}{*}{\shortstack{\textbf{Batch}\\\textbf{Size}}} 
& \multicolumn{4}{ c }{\textbf{Linear Evaluation}} 
& \multicolumn{4}{ c }{\textbf{Fine-tuned}} 
\\
\cline{6-9} \cline{10-13}
\noalign{\smallskip}
& &  & &
&Base LR &WD &\textbf{$S_{lr}$} &\textbf{$S_{wd}$}
&Base LR &WD &\textbf{$S_{lr}$} &\textbf{$S_{wd}$}
\\ 
\Xhline{0.7pt}\noalign{\smallskip}

\multirow{46}{*}{\rotatebox[origin=c]{90}{\small{\ube{}}}}
&\multirow{46}{*}{\shortstack[l]{
Total epochs: 50\\ \\
LR steps: \\
(0, 10, 20, 30) \\\\
LR decay: \\
(1, 0.1, 0.01, 0.001)}}
&\multirow{30}{*}{\rotatebox[origin=c]{90}{ResNet-50}} &\random{}	
&\multirow{20}{*}{128}
&0.025	&\shortstack[l]{\{0.001, 0.01, \\\xspace0.0001, 0.01, 0.01\}}	&\meanstd{1.28}{0.38}	&\meanstd{0.36}{0.06}
&\shortstack[l]{\{0.0025, 0.025, 0.025, \\\xspace0.0025, 0.0025	\}}	
&\shortstack[l]{\{0.0001, 0.001, 0.01, \\\xspace0.01, 0.01\}}	&\meanstd{4.39}{0.75}	&\meanstd{1.54}{0.54}
\\

&&&\insup{}	&	 
&0.025	&\shortstack[l]{\{0.01, 0.0001, \\\xspace0.01, 0.01, 0.01\}}	&\meanstd{10.52}{0.65}	&\meanstd{0.38}{0.26}
&\shortstack[l]{\{0.0025, 0.0025, 0.0025, \\\xspace0.0025, 0.025	\}}	
&\shortstack[l]{\{0.0001, 0.001, 0.001, \\\xspace0.01, 0.0001\}}	&\meanstd{7.35}{1.18}	&\meanstd{2.10}{0.50}
\\	
&&&\moco{}&
&0.025	&\shortstack[l]{\{0.01, 0.01, \\\xspace0.0001, 0.01, 0.01\}}	&\meanstd{3.06}{0.33}	&\meanstd{0.47}{0.18}
&0.0025		
&\shortstack[l]{\{0.0001, 0.01, 0.0001, \\\xspace0.001, 0.0001\}}	&\meanstd{3.94}{0.47}	&\meanstd{0.42}{0.28}
\\
&&&\grid{}&
&0.025	&\shortstack[l]{\{0.0001, 0.0001, \\\xspace0.0001, 0.01, 0.01\}}	&\meanstd{6.91}{0.57}	&\meanstd{0.51}{0.21}
&0.00025	
&\shortstack[l]{\{0.0001, 0.0001, \\\xspace0.0001, 0.01, 0.01\}}	&\meanstd{0.00}{0.00}	&\meanstd{0.64}{0.39}
\\
&&&\virtex{}&
&0.025	&\shortstack[l]{\{0.001, 0.0001, \\\xspace0.01, 0.0001, 0.01\}}	&\meanstd{7.70}{0.45}	&\meanstd{0.41}{0.39}
&0.0025	
&\shortstack[l]{\{0.01, 0.0001, \\\xspace0.001, 0.001, 0.001\}}	&\meanstd{3.76}{0.64}	&\meanstd{0.81}{0.25}
\\
&&&\icmlmatt{}	&
&0.025	&\shortstack[l]{\{0.001, 0.001, \\\xspace0.01, 0.01, 0.0001\}}	&\meanstd{2.20}{0.81}	&\meanstd{0.21}{0.07}
&0.025	
&\shortstack[l]{\{0.01, 0.0001, 0.0001, \\\xspace0.001, 0.0001\}}	&\meanstd{12.38}{1.33}	&\meanstd{0.76}{0.40}
\\
&&&\icmlmtfm{}	&
&0.025	&\shortstack[l]{\{0.01, 0.01, \\\xspace0.01, 0.01, 0.01\}}	&\meanstd{6.08}{1.17}	&\meanstd{0.83}{0.35}
&0.025	
&\shortstack[l]{\{0.01, 0.0001, 0.01, \\\xspace0.01, 0.0001\}}	&\meanstd{6.81}{0.44}	&\meanstd{1.03}{0.46}
\\
&&&\clip{}   &
&0.025	&\shortstack[l]{\{0.01, 0.01, \\\xspace0.01, 0.01, 0.01\}}	&\meanstd{9.88}{0.78}	&\meanstd{0.84}{0.29}
&2.5e-05		&\shortstack[l]{\{0.01, 0.0001, 0.01, \\\xspace0.0001, 0.001\}}	&\meanstd{19.42}{0.79}	&\meanstd{4.36}{4.34}
\\
&&&\visesm{}&
&0.025	&\shortstack[l]{\{0.001, 0.01,\\ \xspace0.01, 0.01, 0.01\}}	&\meanstd{10.04}{0.75}	&\meanstd{0.81}{0.52}
&0.025	
&\shortstack[l]{\{0.01, 0.0001, 0.001, \\\xspace0.001, 0.001\}}	&\meanstd{3.36}{0.87}	&\meanstd{1.84}{0.83}
\\
&&&\viselg{}&
&0.025	&0.01	&\meanstd{14.35}{2.10}	&\meanstd{1.07}{0.24}
&0.0025		
&\shortstack[l]{\{0.001, 0.001, 0.001, \\\xspace0.001, 0.01\}}	&\meanstd{11.24}{2.84}	&\meanstd{1.77}{0.31}
\\

\cline{4-13}\noalign{\smallskip}
&&&\vise{}-123$k$ &\multirow{9}{*}{128}
&-	&-	&-	&-
&0.025	&\shortstack[l]{\{0.01, 0.0001, 0.001, \\\xspace0.001, 0.0001\}}		&\meanstd{3.62}{0.84}		&\meanstd{0.68}{0.39}
\\
&&&\vise{}-308$k$ &	
&-	&-	&-	&-
&0.025	&\shortstack[l]{\{0.001, 0.001, 0.001, \\\xspace0.01, 0.0001\}}		&\meanstd{3.21}{0.87}		&\meanstd{1.07}{0.63}
\\
&&&\vise{}-615$k$ &	
&-	&-	&-	&-
&\shortstack[l]{\{0.0025, 0.0025, 0.025, \\\xspace0.025, 0.025\}}		&\shortstack[l]{\{0.0001, 0.0001, 0.0001, \\\xspace0.001, 0.01\}}		&\meanstd{2.98}{0.97}		&\meanstd{1.08}{0.64}
\\
&&&\visesm{}-$\mathcal{C}$ &	
&-	&-	&-	&-
&\shortstack[l]{\{0.0025, 0.0025, 0.0025, \\\xspace0.025, 0.025	\}}	&\shortstack[l]{\{0.001, 0.0001, 0.0001, \\\xspace0.001, 0.01\}}		&\meanstd{3.00}{0.72}		&\meanstd{1.09}{0.79}
\\
&&&\vise{}-$\mathcal{R}$ &
&-	&-	&-	&-
&\shortstack[l]{\{0.025, 0.0025, 0.025, \\\xspace0.0025, 0.025	\}}	&\shortstack[l]{\{0.001, 0.001, 0.001, \\\xspace0.01, 0.01\}}		&\meanstd{2.18}{0.63}		&\meanstd{0.94}{0.78}
\\
\cline{3-13}\noalign{\smallskip}
&&\multirow{4}{*}{\rotatebox[origin=c]{90}{ResNet-101} }
&\random{}	
&\multirow{5}{*}{64} 
&-	&-	&-	&-
&\shortstack[l]{\{0.0025, 0.00025, 0.00025, \\\xspace0.00025, 0.00025\}}		&\shortstack[l]{\{0.01, 0.0001, 0.001, \\\xspace0.0001, 0.0001\}}		&\meanstd{5.79}{0.67}	&\meanstd{2.00}{1.08}
\\
&&&\insup{}	&		
&-	&-	&-	&-
&0.0025	&\shortstack[l]{\{0.01, 0.0001, 0.001, \\\xspace0.01, 0.0001\}}		&\meanstd{7.76}{2.69}	&\meanstd{1.23}{0.51}
\\
&&&\visesm{} &	
&-	&-	&-	&-
&0.025	&\shortstack[l]{\{0.01, 0.0001, 0.0001, \\\xspace0.0001, 0.0001\}}		&\meanstd{3.16}{0.74}	&\meanstd{1.04}{0.38}
\\
\cline{3-13}\noalign{\smallskip}
&&\multirow{10}{*}{\rotatebox[origin=c]{90}{ResNeXt-101} }
&\random{}	
&\multirow{10}{*}{32} 
&0.025	&0.0001	&\meanstd{2.33}{1.12}	&\meanstd{0.30}{0.13}
&0.025	&\shortstack[l]{\{0.001, 0.0001, 0.01, \\\xspace0.001, 0.001\}}	&\meanstd{2.74}{0.88}	&\meanstd{0.93}{0.36}
\\
&&&\insup{}	&
&0.025	&\shortstack[l]{\{0.0001, 0.0001, 0.0001, \\\xspace0.01, 0.0001\}}	&\meanstd{9.92}{1.04}	&\meanstd{0.12}{0.12}
&0.025	&\shortstack[l]{\{0.01, 0.0001, 0.001, \\\xspace0.01, 0.001\}}	&\meanstd{12.79}{1.48}	&\meanstd{1.37}{0.70}
\\
&&&\wsl{}	&
&0.025	&\shortstack[l]{\{0.0001, 0.0001, 0.01, \\\xspace0.0001, 0.0001\}}	&\meanstd{10.15}{1.42}	&\meanstd{0.19}{0.12}
&0.025	&\shortstack[l]{\{0.001, 0.0001, 0.001, \\\xspace0.001, 0.001\}}	&\meanstd{15.21}{5.96}	&\meanstd{1.25}{0.43}
\\
&&&\visesm{}	&	
&0.025	&\shortstack[l]{\{0.01, 0.001, 0.01, \\\xspace0.01, 0.01\}}	&\meanstd{13.86}{0.80}	&\meanstd{0.32}{0.20}
&0.0025	&\shortstack[l]{\{0.0001, 0.001, 0.0001, \\\xspace0.01, 0.001\}}	&\meanstd{19.07}{8.31}	&\meanstd{1.60}{0.29}
\\
&&&\viselg{}	&	
&0.025	&\shortstack[l]{\{0.0001, 0.01, 0.0001,\\\xspace0.0001, 0.0001\}}	&\meanstd{11.29}{0.79}	&\meanstd{0.12}{0.15}
&0.0025	&\shortstack[l]{\{0.01, 0.001, 0.0001, \\\xspace0.01, 0.0001\}}	&\meanstd{13.73}{0.95}	&\meanstd{0.31}{0.22}
\\

\Xhline{0.7pt}\noalign{\smallskip}
\end{tabular}
}
\caption{
Hyperparameter configurations for best-performing \ube{} models for five random split. Single number are displayed if the configurations are the same across all five experiments.
}
\label{supptab:hp_ube}
\vspace{-0.7cm}
\end{center}
\end{table*}

\begin{table*}
\scriptsize
\begin{center}
\resizebox{\textwidth}{!}{%
\begin{tabular}
{@{\extracolsep{4pt}} l l l ll l l l l l l l l l@{}}
\Xhline{0.7pt}\noalign{\smallskip}
& \multirow{2}{*}{\textbf{\shortstack{Training\\Schedule}}}
& \multirow{2}{*}{\textbf{Backbone}} 
& \multirow{2}{*}{\textbf{Method}}
& \multirow{2}{*}{\textbf{Batch Size}} 
& \multicolumn{4}{ c }{\textbf{Linear Evaluation}} 
& \multicolumn{4}{ c }{\textbf{Fine-tuned}} 
\\
\cline{6-9} \cline{10-13}
\noalign{\smallskip}
& &  & &
&Base LR &WD &\textbf{$S_{lr}$} &\textbf{$S_{wd}$}
&Base LR &WD &\textbf{$S_{lr}$} &\textbf{$S_{wd}$}
\\ 
\Xhline{0.7pt}\noalign{\smallskip}

\multirow{23}{*}{\rotatebox[origin=c]{90}{\small{\politics{}}}}
&\multirow{23}{*}{\shortstack[l]{
Total epochs: 25\\ \\
LR steps: \\
(0, 10, 20) \\\\
LR decay: \\
(1, 0.1, 0.01)}}
&\multirow{15}{*}{ResNet-50} &\random{}	
&\multirow{10}{*}{192}
&0.0025	&0.01	&2.37	&1.49
 & 0.025    & 0.0001 & 0.66 & 0.80 
\\
&&&\insup{}	&	 
&0.025	&0.001	&2.72	&0.07
& 0.0025   & 0.001  & 1.81 & 0.82 
\\	
&&&\moco{}&
&0.025	&0.0001	&1.83	&0.59
 & 0.025    & 0.0001 & 0.75 & 3.46 
\\
&&&\grid{}&
&0.0025	&0.01	&2.02	&0.36
 & 0.00025  & 0.001  & 0.00 & 0.43 
\\
&&&\virtex{}&
&0.025	&0.001	&2.58	&0.44
 & 0.025    & 0.0001 & 0.19 & 2.14 
\\
&&&\icmlmatt{}	&
&0.025	&0.001	&2.37	&0.39
 & 0.025    & 0.0001 & 1.06 & 2.05 
\\
&&&\icmlmtfm{}	&
&0.025	&0.001	&2.90	&0.31
& 0.025    & 0.0001 & 0.69 & 2.48 
\\
&&&\clip{}   &
&0.025	&0.001	&1.54	&0.05
 & 0.000025 & 0.001  & 0.45 & 0.24 
\\
&&&\visesm{}&
&0.025	&0.001	&2.60	&0.27
& 0.025    & 0.0001 & 0.61 & 2.23 
\\
&&&\viselg{}&
&0.025	&0.001	&2.69	&0.34
& 0.00025  & 0.01   & 3.83 & 0.15
\\
\cline{4-13}\noalign{\smallskip}
&&&\vise{}-123$k$ &\multirow{5}{*}{192}
&-	&-	&-	&-
 & 0.025 & 0.0001 & 0.39 & 0.11 
\\
&&&\vise{}-308$k$ &	
&-	&-	&-	&-
& 0.025 & 0.0001 & 1.01 & 2.41 
\\
&&&\vise{}-615$k$ &	
&-	&-	&-	&-
& 0.025 & 0.0001 & 0.72 & 1.71 
\\
&&&\visesm{}-$\mathcal{C}$ &	
&-	&-	&-	&-
& 0.025 & 0.0001 & 0.42 & 2.54 
\\
&&&\vise{}-$\mathcal{R}$ &
&-	&-	&-	&-
& 0.025 & 0.0001 & 0.36 & 2.17 
\\
\cline{3-13}\noalign{\smallskip}
&&\multirow{3}{*}{ResNet-101} 
&\random{}	
&\multirow{3}{*}{192} 
&-	&-	&-	&-
& 0.025 & 0.001    & 0.48   & 0.69 
\\
&&&\insup{}	&		
&-	&-	&-	&-
& 0.025  & 0.0001 & 0.61   & 0.77 
 
\\
&&&\visesm{} &	
&-	&-	&-	&-
& 0.025  & 0.0001 & 0.49   & 2.16 
\\
\cline{3-13}\noalign{\smallskip}
&&\multirow{5}{*}{ResNeXt-101} 
&\random{}	
&\multirow{5}{*}{192} 
&0.0025	&0.001	&0.07	&0.05
 & 0.025  & 0.0001 & 0.61   & 0.80 
\\
&&&\insup{}	&
&0.0025	&0.0001	&0.08	&0.06
 & 0.025  & 0.0001 & 1.26   & 1.76 
\\
&&&\wsl{}	&
&0.0025	&0.001	&3.36	&0.01
& 0.0025 & 0.0001 & 2.63   & 1.37 
\\
&&&\visesm{}	&	
&0.025	&0.0001	&0.59	&0.48
& 0.025  & 0.0001 & 0.80   & 2.74 
\\
&&&\viselg{}	&	
&0.025	&0.001	&0.64	&0.53
 & 0.025  & 0.0001 & 0.16   & 2.03 

\\
\Xhline{0.7pt}\noalign{\smallskip}
\end{tabular}
}
\caption{
Hyperparameter configurations for best-performing \politics{} models.
``Batch Size'' presents the total mini batch size across 8GPUs. For fine-tuned settings, some learning processes are stopped early.
}
\label{supptab:hp_politics}
\vspace{-0.7cm}
\end{center}
\end{table*}

\begin{table*}
\scriptsize
\begin{center}
\resizebox{\textwidth}{!}{%
\begin{tabular}
{@{\extracolsep{4pt}} l l l ll l l l l l l l l l@{}}
\Xhline{0.7pt}\noalign{\smallskip}
& \multirow{2}{*}{\textbf{\shortstack{Training\\Schedule}}}
& \multirow{2}{*}{\textbf{Backbone}} 
& \multirow{2}{*}{\textbf{Method}}
& \multirow{2}{*}{\textbf{Batch Size}} 
& \multicolumn{4}{ c }{\textbf{Linear Evaluation}} 
& \multicolumn{4}{ c }{\textbf{Fine-tuned}} 
\\
\cline{6-9} \cline{10-13}
\noalign{\smallskip}
& &  & &
&Base LR &WD &\textbf{$S_{lr}$} &\textbf{$S_{wd}$}
&Base LR &WD &\textbf{$S_{lr}$} &\textbf{$S_{wd}$}
\\ 
\Xhline{0.7pt}\noalign{\smallskip}

\multirow{23}{*}{\rotatebox[origin=c]{90}{\small{\hm{}}}}
&\multirow{23}{*}{\shortstack[l]{
Total epochs: 30\\ \\
LR steps: \\
(0, 20) \\\\
LR decay: \\
(1, 0.5)}}
&\multirow{15}{*}{ResNet-50} &\random{}	
&\multirow{10}{*}{64}
&0.025	&0.01	&0.0184	&0.0022
&0.025	&0.01	&0.0322	&0.0014
\\
&&&\insup{}	&	 
&0.025	&0.01	&0.0184	&0.0006
&0.025	&0.0001	&0.0346	&0.0064
\\	
&&&\moco{}&
&0.025	&0.01 &0.0364	&0.0009
&0.025	&0.001	&0.0265	&0.0022
\\
&&&\grid{}&
&0.025	&0.01 &0.0265	&0.001
&0.025	&0.0001	&0.0442	&0.0426
\\
&&&\virtex{}&
&0.025	&0.01	&0.0194	&0.0018
&0.025	&0.001	&0.0301	&0.0011
\\
&&&\icmlmatt{}	&
&0.025	&0.01	&0.018	&0.0002
&0.025	&0.001	&0.0353	&0.0011
\\
&&&\icmlmtfm{}	&
&0.025	&0.01	&0.0217	&0.002
&0.025	&0.01	&0.0395	&0.0024
\\
&&&\clip{}   &
&0.025	&0.01	&0.0583	&0.0007
&0.00025	&0.01	&0	&0.0021
\\
&&&\visesm{}&
&0.025	&0.01	&0.0453	&0.0015
&0.025	&0.01	&0.0403	&0.0061
\\
&&&\viselg{}&
&0.025	&0.01	&0.0191	&0.0018
&0.025	&0.0001	&0.0284	&0.0063
\\
\cline{4-13}\noalign{\smallskip}
&&&\vise{}-123$k$ &\multirow{5}{*}{64}
&-	&-	&-	&-
&0.025	&0.01	&0.0452	&0.0033
\\
&&&\vise{}-308$k$ &	
&-	&-	&-	&-
&0.025	&0.01	&0.0324	&0.0014
\\
&&&\vise{}-615$k$ &	
&-	&-	&-	&-
&0.025	&0.0001	&0.0341	&0.0074
\\
&&&\visesm{}-$\mathcal{C}$ &	
&-	&-	&-	&-
&0.025	&0.001	&0.029	&0.0033
\\
&&&\vise{}-$\mathcal{R}$ &
&-	&-	&-	&-
&0.025	&0.001	&0.0318	&0.0043
\\
\cline{3-13}\noalign{\smallskip}
&&\multirow{3}{*}{ResNet-101} 
&\random{}	
&\multirow{3}{*}{32} 
&-	&-	&-	&-
&0.025	&0.01	&0.0404	&0.0062
\\
&&&\insup{}	&		
&-	&-	&-	&-
&0.025	&0.01	&0.0378	&0.0053
\\
&&&\visesm{} &	
&-	&-	&-	&-
&0.025	&0.01	&0.0368	&0.0088
\\
\cline{3-13}\noalign{\smallskip}
&&\multirow{5}{*}{ResNeXt-101} 
&\random{}	
&\multirow{5}{*}{16} 
&0.00025	&0.0001	&0.0046	&0
&0.025	&0.0001	&0.0365	&0.0015
\\
&&&\insup{}	&
&0.025	&0.01	&0.02	&0.0006
&0.025	&0.001	&0.0348	&0.009
\\
&&&\wsl{}	&
&0.025	&0.01	&0.0308	&0.001
&0.025	&0.01	&0.032	&0.0018
\\
&&&\visesm{}	&	
&0.025	&0.01	&0.0273	&0.0016
&0.025	&0.01	&0.0404	&0.0021
\\
&&&\viselg{}	&	
&0.025	&0.01	&0.0161	&0.0018
&0.025	&0.01	&0.0324	&0.0051
\\
\Xhline{0.7pt}\noalign{\smallskip}
\end{tabular}
}
\caption{
Hyperparameter configurations for best-performing \hm{} models. The text encoder is used as a feature extractor in these experiments.
}
\label{supptab:hp_hm}
\vspace{-0.5cm}
\end{center}
\end{table*}

\begin{table*}
\scriptsize
\begin{center}
\resizebox{\textwidth}{!}{%
\begin{tabular}
{@{\extracolsep{4pt}} l l l ll l l l l l l l l l@{}}
\Xhline{0.7pt}\noalign{\smallskip}
& \multirow{2}{*}{\textbf{\shortstack{Training\\Schedule}}}
& \multirow{2}{*}{\textbf{Backbone}} 
& \multirow{2}{*}{\textbf{Method}}
& \multirow{2}{*}{\textbf{Batch Size}} 
& \multicolumn{4}{ c }{\textbf{Linear Evaluation}} 
& \multicolumn{4}{ c }{\textbf{Fine-tuned}} 
\\
\cline{6-9} \cline{10-13}
\noalign{\smallskip}
& &  & &
&Base LR &WD &\textbf{$S_{lr}$} &\textbf{$S_{wd}$}
&Base LR &WD &\textbf{$S_{lr}$} &\textbf{$S_{wd}$}
\\ 
\Xhline{0.7pt}\noalign{\smallskip}

\multirow{9}{*}{\rotatebox[origin=c]{90}{\small{\cub{}}}}
&\multirow{9}{*}{\shortstack[l]{
Total epochs: 300 \\ \\
LR steps: \\
(0, 100, 200) \\\\
LR decay: \\
(1, 0.1, 0.01)}}
&\multirow{4}{*}{ResNet-50} &\random{}	
&\multirow{4}{*}{128}
&0.025	&0.001	&1.17	&0.20
&0.025	&0.01	&23.12	&14.49
\\
&&&\insup{}	&	 
&0.025	&0.001	&25.25	&0.62
&0.025	&0.01	&10.14	&0.99
\\
&&&\visesm{}		&	
&0.025	&0.0001	&4.04	&0.75
&0.025	&0.001	&10.41	&1.31
\\
&&&\viselg{}	&		
&0.025	&0.0001	&3.87	&0.94
&0.025	&0.001	&4.71	&0.80
\\
\cline{3-13}\noalign{\smallskip}
&&\multirow{5}{*}{ResNeXt-101} 
&\random{}	
&\multirow{5}{*}{32} 
&0.025	&0.0001	&1.86	&0.91
&0.025	&0.01	&27.50	&8.04
\\
&&&\insup{}	&
&0.025	&0.0001	&8.83	&0.06
&0.025	&0.0001	&2.97	&3.23
\\
&&&\wsl{}	&
&0.025	&0.0001	&23.98	&0.72
&0.0025	&0.001	&1.13	&0.00
\\

&&&\visesm{}	&	
&0.025	&0.001	&3.28	&0.96
&0.025	&0.001	&29.39	&0.60
\\
&&&\viselg{}	&	
&0.025	&0.0001	&3.19	&1.22
&0.025	&0.001	&30.14	&2.42
\\

\Xhline{0.7pt}\noalign{\smallskip}
\end{tabular}
}
\caption{
Hyperparameter configurations for best-performing \cub{} models.
}
\label{supptab:hp_cub}
\vspace{-0.2cm}
\end{center}
\end{table*}

\begin{table*}
\scriptsize
\begin{center}
\resizebox{0.8\textwidth}{!}{%
\begin{tabular}
{@{\extracolsep{4pt}} l l l ll l l l l l@{}}
\Xhline{0.7pt}\noalign{\smallskip}
& \multirow{2}{*}{\textbf{\shortstack{Training\\Schedule}}}
& \multirow{2}{*}{\textbf{Backbone}} 
& \multirow{2}{*}{\textbf{Method}}
& \multirow{2}{*}{\textbf{Batch Size}} 
& \multicolumn{4}{ c }{\textbf{Image + Text (Fine-tuned)}} 
\\
\cline{6-9}
\noalign{\smallskip}
& &  & &
&Base LR &WD &\textbf{$S_{lr}$} &\textbf{$S_{wd}$}
\\ 
\Xhline{0.7pt}\noalign{\smallskip}

\multirow{12}{*}{\rotatebox[origin=c]{90}{\small{\hm{}}}}
&\multirow{12}{*}{\shortstack[l]{
Total epochs: 30\\ \\
LR steps: \\
(0, 20) \\\\
LR decay: \\
(1, 0.5)}}
&\multirow{10}{*}{ResNet-50} &\random{}	
&\multirow{10}{*}{64}
& 0.00025   & 0.001     & 0.04   & 0.0099 \\
&&&\insup{}	&	 
& 0.00025   & 0.001     & 0.0613 & 0.0052 \\
&&&\moco{}&
& 0.00025   & 0.01      & 0.0668 & 0.0005 \\
&&&\grid{}&
& 0.00025   & 0.001     & 0.0497 & 0.0067 \\
&&&\virtex{}&
& 0.00025   & 0.001     & 0.0704 & 0.037  \\
&&&\icmlmatt{}	&
& 0.00025   & 0.01      & 0.0684 & 0.0029 \\
&&&\icmlmtfm{}	&
& 0.00025   & 0.0001    & 0.0713 & 0.0073 \\
&&&\clip{}   &
& 0.00025   & 0.001     & 0      & 0.0026 \\

&&&\visesm{}&
& 0.00025   & 0.0001    & 0.0662 & 0.0039 \\
&&&\viselg{}&
& 0.00025   & 0.01      & 0.0697 & 0.0045 \\
\cline{3-9}\noalign{\smallskip}
&&\multirow{2}{*}{ResNeXt-101} 
&\wsl{}
&\multirow{2}{*}{16} 
 & 0.00025   & 0.001     & 0.0628 & 0.0022 \\

&&&\viselg{}	&	
& 0.00025   & 0.01      & 0.0716 & 0.0052\\
\Xhline{0.7pt}\noalign{\smallskip}
\end{tabular}
}
\caption{
Hyperparameter configurations for best-performing \hm{} models: Image + Text (Fine-tuned).
}
\label{supptab:hp_mm_full}
\end{center}
\end{table*}

\begin{table*}[!htbp]
\scriptsize
\begin{center}
\resizebox{0.8\textwidth}{!}{%
\begin{tabular}
{@{\extracolsep{4pt}} l l l ll l l l l l@{}}
\Xhline{0.7pt}\noalign{\smallskip}
& \multirow{2}{*}{\textbf{\shortstack{Training\\Schedule}}}
& \multirow{2}{*}{\textbf{Backbone}} 
& \multirow{2}{*}{\textbf{Method}}
& \multirow{2}{*}{\textbf{Batch Size}} 
& \multicolumn{4}{ c }{\textbf{Image Only (Fine-tuned)}} 
\\
\cline{6-9}
\noalign{\smallskip}
& &  & &
&Base LR &WD &\textbf{$S_{lr}$} &\textbf{$S_{wd}$}
\\ 
\Xhline{0.7pt}\noalign{\smallskip}

\multirow{12}{*}{\rotatebox[origin=c]{90}{\small{\hm{}}}}
&\multirow{12}{*}{\shortstack[l]{
Total epochs: 30\\ \\
LR steps: \\
(0, 20) \\\\
LR decay: \\
(1, 0.5)}}
&\multirow{10}{*}{ResNet-50} &\random{}	
&\multirow{10}{*}{64}
& 0.0025  & 0.0001 & 0.0058 & 0.0043 \\

&&&\insup{}	&	 
& 0.0025  & 0.001  & 0.0206 & 0.0099 \\

&&&\moco{}&
& 0.025   & 0.01   & 0.0117 & 0.0015 \\

&&&\grid{}&
& 0.00025 & 0.001  & 0      & 0.0154 \\
 
&&&\virtex{}&
& 0.025   & 0.0001 & 0.007  & 0.0171 \\
 
&&&\icmlmatt{}	&
& 0.025   & 0.01   & 0.01   & 0.0186 \\
 
&&&\icmlmtfm{}	&
& 0.025   & 0.001  & 0.0146 & 0.0224 \\
 
&&&\clip{}   &
& 0.00025 & 0.001  & 0      & 0      \\
 
&&&\visesm{}&
& 0.025   & 0.0001 & 0.0168 & 0.0044 \\

&&&\viselg{}&
& 0.0025  & 0.01   & 0.0185 & 0.0146 \\
\cline{3-9}\noalign{\smallskip}
&&\multirow{2}{*}{ResNeXt-101} 
&\wsl{}
&\multirow{2}{*}{16} 
& 0.00025 & 0.001 & 0.0146 & 0.0023 \\

&&&\viselg{}	&	
& 0.0025  & 0.001 & 0.0161 & 0.0122 \\

\Xhline{0.7pt}\noalign{\smallskip}
\end{tabular}
}
\caption{
Hyperparameter configurations for best-performing \hm{} models: Image Only (Fine-tuned).
}
\label{supptab:hp_mm_img}
\end{center}
\end{table*}

\begin{table*}[!htbp]
\scriptsize
\begin{center}
\resizebox{0.8\textwidth}{!}{%
\begin{tabular}
{@{\extracolsep{4pt}} l l l ll l l l l l@{}}
\Xhline{0.7pt}\noalign{\smallskip}
& \multirow{2}{*}{\textbf{\shortstack{Training\\Schedule}}}
& \multirow{2}{*}{\textbf{Backbone}} 
& \multirow{2}{*}{\textbf{Method}}
& \multirow{2}{*}{\textbf{Batch Size}} 
& \multicolumn{4}{ c }{\textbf{Image Only (Linear)}} 
\\
\cline{6-9}
\noalign{\smallskip}
& &  & &
&Base LR &WD &\textbf{$S_{lr}$} &\textbf{$S_{wd}$}
\\ 
\Xhline{0.7pt}\noalign{\smallskip}

\multirow{12}{*}{\rotatebox[origin=c]{90}{\small{\hm{}}}}
&\multirow{12}{*}{\shortstack[l]{
Total epochs: 30\\ \\
LR steps: \\
(0, 20) \\\\
LR decay: \\
(1, 0.5)}}
&\multirow{10}{*}{ResNet-50} &\random{}	
&\multirow{10}{*}{64}
& 0.0025  & 0.0001 & 0.0075 & 0      \\
 
&&&\insup{}	&	 
& 0.025   & 0.01   & 0.0055 & 0.0002 \\
 
&&&\moco{}&
& 0.025   & 0.0001 & 0.0157 & 0.0004 \\
 
&&&\grid{}&
& 0.00025 & 0.0001 & 0.0125 & 0.0004 \\
 
&&&\virtex{}&
& 0.025   & 0.01   & 0.0078 & 0.0007 \\
  
&&&\icmlmatt{}	&
& 0.025   & 0.001  & 0.006  & 0.0001 \\
 
&&&\icmlmtfm{}	&
& 0.025   & 0.01   & 0.0179 & 0.0001 \\
 
&&&\clip{}   &
& 0.025   & 0.01   & 0.043  & 0.0004 \\
 
&&&\visesm{}&
& 0.0025  & 0.0001 & 0.0089 & 0      \\

&&&\viselg{}&
 & 0.025   & 0.01   & 0.0291 & 0.0009
\\
\cline{3-9}\noalign{\smallskip}
&&\multirow{2}{*}{ResNeXt-101} 
&\wsl{}
&\multirow{2}{*}{16} 
& 0.025 & 0.01 & 0.0092 & 0.0004 \\

&&&\viselg{}	&	
& 0.025 & 0.01 & 0.0299 & 0.0018 \\

\Xhline{0.7pt}\noalign{\smallskip}
\end{tabular}
}
\caption{
Hyperparameter configurations for best-performing \hm{} models: Image Only (Linear).
}
\label{supptab:hp_mm_imgfz}
\end{center}
\end{table*}

\subsection{Other Pre-training Methods}
We use the publicly available pre-trained models for other compared baseline methods\footnote{Links for the publicly available pre-trained models: 
\href{https://github.com/facebookresearch/WSL-Images}{\wsl{}},
\href{https://github.com/facebookresearch/moco}{\moco{}},
\href{https://github.com/facebookresearch/grid-feats-vqa}{\grid{}},
\href{https://github.com/kdexd/virtex}{\virtex{}},
\href{https://europe.naverlabs.com/research/computer-vision/icmlm/}{ \icmlm{}},
\href{https://github.com/OpenAI/CLIP}{\clip}.
}
except for ImageNet pretraining with ResNeXt-101 backbone. We train that model with 100 epochs with learning rate decay schedule of $(30, 60, 90)$ and scaling factor of 0.1.
Note that the pre-trained model for \clip{} adopts a modified ResNet-50 architecture. See~\cite{radford2021learning} for details.

\subsection{Downstreaming Tasks Setup}
\label{supsubsec:ft_details}
\paragraph{Tasks summary}
The statistics of these tasks and the associated datasets are listed in Table~\ref{tab:datasets}.

\paragraph{Implementation}
Similar to the pre-training models, we use Pytorch and NVIDIA Tesla V100 16GB GPUs for the transfer learning experiments. Table~\ref{supptab:infra} summarizes other implementation details including average runtime. The same data augmentation are employed as the pretraining stage.
To encode raw text of the multi-modal experiments, we use RoBERTa base from fairseq~\cite{ott2019fairseq}\footnote{
\href{https://github.com/pytorch/fairseq/blob/master/examples/roberta/README.md}{Link for the publicly available pre-trained RoBERTa-base model}}.

\paragraph{Optimization and training details}
We use stochastic gradient descent with 0.9 momentum for image only models and Adam optimization with decoupled weight decay~\cite{loshchilov2018decoupled} for multi-modal experiments.
Following~\cite{wslimageseccv2018}, we conduct a coarse grid search to find the learning rate and weight decay values using \val{} split.
The learning rate is set as $\text{Base LR} / 256 \times \text{batchsize}$, where Base LR is chosen from $\{0.025, 0.0025, 0.00025\}$. 
For pre-training method \clip{}, we expand the search to $\{0.025, 0.0025, 0.00025, 0.000025, 0.0000025\}$.
The bound for weight decay is:  $\{0.01, 0.001, 0.0001\}$.
We also report the model performance sensitivity to learning rate ($S_{lr}$) and weight decay ($S_{wd}$) values. $S_{lr}$ is defined as the standard deviation of the model performance across the range of learning rate considered given the optimal weight decay value. 
Similarly, $S_{wd}$ is the standard deviation across the range of weight decay values given the optimal learning rate. 
Tables~\ref{supptab:hp_ube}-\ref{supptab:hp_mm_imgfz} show the training details and hyperparameter configurations of all the experiments in the main text.

\newpage
{\small
\bibliographystyle{ieee_fullname}
\bibliography{egbib}
}

\end{document}